\documentclass[pmlr, usenames, dvipsnames]{jmlr}

\usepackage{longtable}

 \usepackage{booktabs}
 \usepackage{multirow}
 \usepackage{enumerate}
\usepackage{arydshln}
\usepackage{microtype}  

\usepackage[load-configurations=version-1]{siunitx} 

\usepackage{bm}
\usepackage{physics}
\usepackage{algorithm}
\usepackage{algpseudocode}
\usepackage{mathtools}
\usepackage{colortbl}
\usepackage{xurl}

\definecolor{darkblue}{rgb}{0.0, 0.0, 0.55}

\algrenewcommand\algorithmicrequire{\textbf{Input:}}
\algrenewcommand\algorithmicensure{\textbf{Output:}}
\algnewcommand\algorithmicforeach{\textbf{for each}}
\algdef{S}[FOR]{ForEach}[1]{\algorithmicforeach\ #1\ \algorithmicdo}

\usepackage[acronym, toc]{glossaries}
\makeglossaries

\newcommand{\Rom}[1]{\uppercase\expandafter{\romannumeral#1\relax}}

\newacronym{ml}{ML}{machine learning}
\newacronym{ai}{AI}{artificial intelligence}
\newacronym{dl}{DL}{deep learning}
\newacronym{ad}{AD}{automatic differentiation}
\newacronym{ehr}{EHR}{electronic health record}
\newacronym{nhs}{NHS}{National Health Service}
\newacronym{mimic3}{MIMIC-\Rom{3}}{Medical Information Mart for Intensive Care ~\Rom{3}}
\newacronym{mimic4}{MIMIC-\Rom{4}}{Medical Information Mart for Intensive Care ~\Rom{4}}
\newacronym{wsic}{WSIC}{Whole Systems Integrated Care}
\newacronym{icare}{ICARE}{Imperial's Clinical Analytics, Research and Evaluation}
\newacronym{gp}{GP}{general practitioner}
\newacronym{rnn}{RNN}{recurrent neural network}
\newacronym{gru}{GRU}{gated recurrent unit}
\newacronym{rtd}{RtD}{Routes to Diagnosis}
\newacronym{ode}{ODE}{ordinary differential equation}
\newacronym{sde}{SDE}{stochastic differential equation}
\newacronym{phe}{PHE}{Public Health England}
\newacronym{dag}{DAG}{directed acyclic graph}
\newacronym{icht}{ICHT}{Imperial College Health Trust}
\newacronym{eda}{EDA}{exploratory data analysis}
\newacronym{nlp}{NLP}{natural language processing}
\newacronym{mlp}{MLP}{multilayer perceptron}
\newacronym{sgd}{SGD}{stochastic gradient descent}
\newacronym{ivp}{IVP}{initial value problem}
\newacronym{iqr}{IQR}{interquartile range}
\newacronym{api}{API}{application programming interface}
\newacronym{gpu}{GPU}{graphical processing unit}
\newacronym{icu}{ICU}{intensive care unit}
\newacronym{mhra}{MHRA}{Medicines and Healthcare products Regulatory Agency}
\newacronym{svm}{SVM}{support vector machine}
\newacronym{rf}{RF}{random forest}
\newacronym{nb}{NB}{naive Bayes}
\newacronym{gnn}{GNN}{graph neural network}
\newacronym{dpm}{DPM}{disease progression modeling}
\newacronym{bidmc}{BIDMC}{Beth Israel Deaconess Medical Center}
\newacronym{ccs}{CCS}{Clinical Classifications Software}
\newacronym{auc}{AUC}{area under the receiver operating characteristic curve}

\makeatletter
\def\set@curr@file#1{\def\@curr@file{#1}} 
\makeatother

\theorembodyfont{\upshape}
\theoremheaderfont{\scshape}
\theorempostheader{:}
\theoremsep{\newline}

\jmlryear{2022}

\title[\texttt{ICE-NODE}]{\texttt{ICE-NODE}: Integration of Clinical Embeddings \\ with  Neural Ordinary Differential Equations}

\author{\Name{Asem Alaa}
      \Email{asem.a.abdelaziz@imperial.ac.uk}\\ 
      \addr Departments of Computing, Mathematics, and Cancer and Surgery\\ \& UKRI Centre for Doctoral Training in AI for Healthcare, \\ 
      Imperial College London, UK 
      \AND
      \Name{Erik Mayer}
      \Email{e.mayer@imperial.ac.uk}\\ 
      \addr Department of Cancer and Surgery, \\ Imperial College London, UK 
      \AND
      \Name{Mauricio Barahona}
      \Email{m.barahona@imperial.ac.uk}\\ 
      \addr Department of Mathematics \\
      \& Centre for Mathematics of Precision Healthcare, \\ 
      Imperial College London, UK
      }

\begin{document}

\maketitle

\begin{abstract}
  Early diagnosis of disease can lead to improved health outcomes, including higher survival rates and lower treatment costs. With the massive amount of information available in \glspl*{ehr}, there is great potential to use \gls*{ml} methods to model disease progression aimed at early prediction of disease onset and other outcomes. In this work, we employ recent innovations in neural \acrshortpl*{ode} combined with rich semantic embeddings of clinical codes to harness the full temporal information of \glspl*{ehr}. We propose \texttt{ICE-NODE} (Integration of Clinical Embeddings with Neural Ordinary Differential Equations), an architecture that temporally integrates embeddings of clinical codes and neural \acrshortpl*{ode} to learn and predict patient trajectories in \glspl*{ehr}. We apply our method to the publicly available \acrshort*{mimic3} and \acrshort*{mimic4} datasets, and we find improved prediction results compared to state-of-the-art methods, specifically for clinical codes that are not frequently observed in \glspl*{ehr}. We also show that  \texttt{ICE-NODE} is more competent at predicting certain medical conditions, like acute renal failure, pulmonary heart disease and birth-related problems, where the full temporal information could provide important information.   Furthermore, \texttt{ICE-NODE} is also able to produce patient risk trajectories over time that can be exploited for further detailed predictions of disease evolution. 
\end{abstract}

\section{Introduction}






With the wider availability of \glspl*{ehr}, and the massive amount of information they contain, there is a rising demand to exploit such data using current advances in \gls*{ml} to improve healthcare outcomes. For instance, many healthcare systems across the world
suffer from delayed cancer diagnosis, leading to lowered survival rates in cancer patients ~\citep{arnold2019progress}. In 2020, over 19 million people were diagnosed with cancer and around 10 million people died from cancer~\citep{ferlay2021cancer}. In England and Wales alone, over three hundred and fifty thousand cancer cases are diagnosed yearly (averaged over 2016-2018), and from those diagnosed with cancer in 2010-2011 only 50\% survive for ten years or more~\citep{CanceResearchUK2015}. 
Employing new innovations in \gls*{ml} for \gls*{dpm} has the potential to take advantage of the available data in patient histories to aid clinicians in their quest for early prediction of disease onset.

Information in \glspl*{ehr} is stored as timestamped \emph{medical concepts} of diverse type (e.g., symptoms, procedures, lab tests). 
The medical concepts in \glspl*{ehr} are 
stored 
using coding schemes, such as \emph{SNOMED CT}\footnote{\url{https://termbrowser.nhs.uk/}} or \emph{ICD}\footnote{\url{https://www.who.int/classifications/classification-of-diseases}},
which provide a comprehensive, interrelated classification of diseases within a compactly structured data system.

\begin{figure}[t]
    \centering
    \includegraphics[width=.9\linewidth]{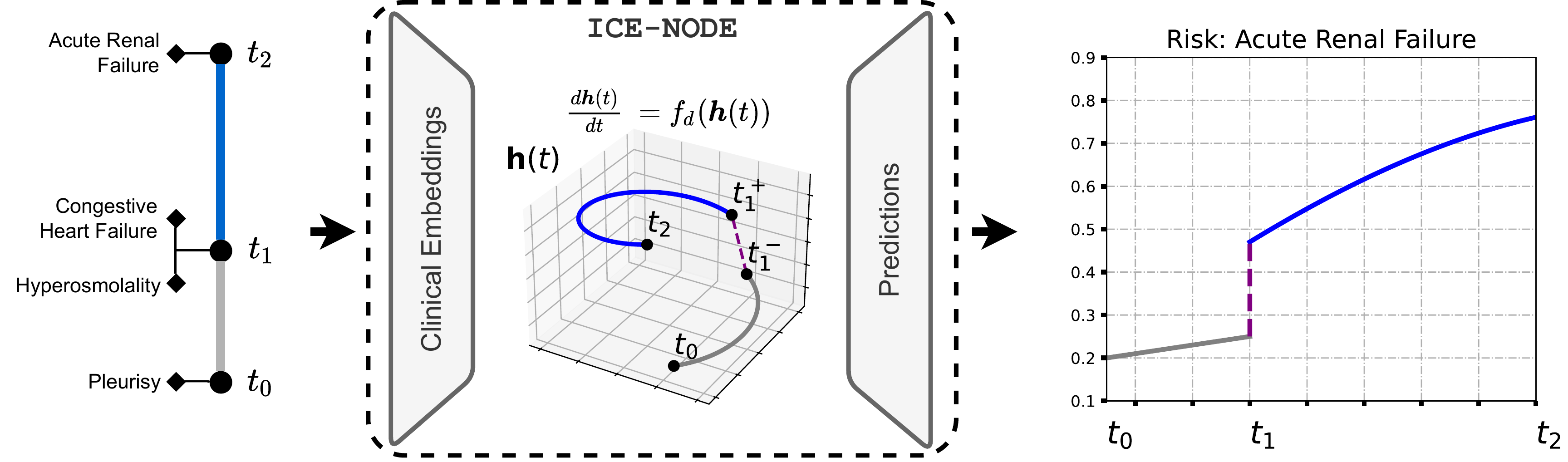}
    \caption{\footnotesize  A schematic view of \texttt{ICE-NODE}. The input is an \gls*{ehr} at three consecutive timestamps $\{t_0, t_1, t_2\}$
    \texttt{ICE-NODE} learns a neural \acrshort*{ode} model from 
    the clinical codes at each timestamp by generating predictions at the observation times of the samples in the training set. Once the model is learnt, 
    it can generate a health state trajectory over time and continuous-time clinical embeddings that can be used to predict clinical codes, enabling risk predictions for medical conditions over time (e.g., for `Acute renal failure' in this case).}  
    \label{fig:ehr}
\end{figure}

The complexity of the information in \glspl*{ehr} poses a number of challenges for many \gls*{ml} predictive models, which usually make strong assumptions about the data when dealing with feature extraction and representation of clinical codes, and with the treatment of the temporal dimension in the patient history. Many \gls*{dpm} methods \citep[e.g.][]{Cherry2020ACancer.,  Weegar2020UsingRepresentations} transform the \gls*{ehr} into a tabular format discarding the temporal information altogether, whereas other approaches \citep[e.g.][]{Choi2018MiME:Healthcare, Choi2016RETAIN:Mechanism} only capture the \emph{sequential} information within the \gls*{ehr}. \glspl*{rnn} and other \gls*{nlp}-inspired models usually underpin the methods of this latter group. Yet, while these methods can aid in capturing the temporal ordering of medical concepts, they do not capture the irregularity of time intervals between consecutive timestamps. Such assumptions may limit the learning capabilities of models applied to such temporal patterns, which might be important for particular medical conditions.
%


Recently, a new set of \gls*{ml} methods \citep{NEURIPS2018_69386f6b, Brouwer2019GRU-ODE-Bayes:Series, Rubanova2019LatentSeries, Kidger2020NeuralSeries, Kidger2022} have developed a class of neural networks that include \glspl*{ode} through implicit layers, as opposed to the common feed-forward static layers. Such neural \glspl*{ode} models have the potential to learn the dynamics of \glspl*{ehr} as sporadically observed time series, irregularly sampled in time and feature dimensions. Indeed, neural \glspl*{ode} have shown superiority over other methods (including different variants of GRU and LSTM models) when applied to such time series~\citep{Brouwer2019GRU-ODE-Bayes:Series}.

In this work, we aim to bring advanced \gls*{ml} methods apt for the analysis of sporadically observed time series to the disease progression modeling task. Specifically, we propose \texttt{ICE-NODE},   
a model that learns from the timestamped sequences of diagnostic codes in  patient trajectories as they interact with healthcare providers. \texttt{ICE-NODE} takes into account explicitly the temporal dimension in the \glspl*{ehr} using neural \glspl*{ode}, with a flexible framework that enables integrating advanced embedding modules for the medical concepts. 
Below, we present, test, and analyse \texttt{ICE-NODE} with two different integrated modules for the embedding of medical concepts: simple matrix embeddings and \texttt{GRAM} embeddings~\citep{Choi2017GRAM:Learning}.

%
\texttt{ICE-NODE} achieves high predictive accuracy on unseen patient trajectories, outperforming the state-of-the-art methods in certain \gls*{dpm} tasks. 
In particular, our analysis with \texttt{ICE-NODE} identifies a set of diagnostic codes, including acute renal failure and pulmonary heart disease, which are predicted with a significantly higher accuracy when the irregularity of time intervals between timestamps is incorporated explicitly. 
Furthermore, we show that \texttt{ICE-NODE} enables access to an inherent (hidden) temporal variable that describes the patient state at any time between timestamps in the patient history. This feature can be exploited to generate a time-continuous disease risk for the patient that could be used as an aid in clinical decision-making (see Fig.~\ref{fig:ehr} for an illustration). Studying and visualising the time-continuous trajectories of multiple medical conditions in this way can enable better understanding of the patient health state and the temporal interlinkages between conditions.

We apply our methodology to two publicly available, de-identified, \glspl*{ehr} datasets available through \emph{PhysioNet}~\citep{goldberger2000physiobank}: \acrshort*{mimic3}~\citep{mimic3, johnson2016mimic}, which contains \glspl*{ehr} for more than 46K patients over an 11-year period, and \acrshort*{mimic4}~\citep{mimic4}, which contains \glspl*{ehr} for more than 256K patients over 11 years. The source code of our \texttt{ICE-NODE} implementation, along with the numerical experiments in this paper, is available at \url{https://github.com/barahona-research-group/ICE-NODE}.

\subsection*{Generalisable Insights about Machine Learning in the Context of Healthcare}


Medical conditions evolve in time and may interact with each other, and such complex dynamical interactions are reflected in \gls*{ehr} patient histories. 
%
The main motivation behind developing \texttt{ICE-NODE} is to deploy recent innovations in neural \glspl*{ode} that allow us to take into account the detailed temporal information and the full complexity of diagnostic codes to learn the underlying dynamical health state of patients from timestamped diagnostic codes collected through~\glspl*{ehr}.
%
In summary:
\begin{itemize}
    \item We develop a framework that represents timestamped medical diagnostic codes from patient histories and feeds them to a specialised neural \gls*{ode} model to learn the temporal evolution of medical conditions. \texttt{ICE-NODE} incorporates the full temporal information of \glspl*{ehr}, unlike traditional methods that ignore time or use only sequential orderings. 
    \item We show through performance analysis that exploiting the full temporal information results in a significant improvement in the prediction accuracy of disease onset for particular medical conditions like acute renal failure and pulmonary heart disease.
    \item We learn time-continuous trajectories for disease risk prediction, and discuss opportunities to exploit these trajectories for a better understanding of patient health.  
\end{itemize}


\section{Related Work}


The high explanatory potential of \gls*{ehr} data and their availability through databases such as \emph{PhysioNet}~\citep{goldberger2000physiobank} has motivated a large body of research on learning from \glspl*{ehr}, while tackling their complex and unstructured format \citep{Xiao2018OpportunitiesReview,Choi2016Multi-layerConcepts, Choi2016RETAIN:Mechanism,Choi2017GRAM:Learning,Cai2018MedicalAttention,Choi2018MiME:Healthcare,Zhang2018Patient2Vec:Record, Peng2019TemporalEmbedding,Wang2019Inpatient2Vec:Inpatients,Zhu2019GraphDisease,Choi2020LearningTransformer,Xu2020EHR2Vec:Mechanism}.  

Many of the traditional \gls*{dpm} methods in the literature transform the unstructured \glspl*{ehr} into a tabular format before applying common machine learning methods~\citep{Cherry2020ACancer., Ferroni2019BreastApproach, Weegar2020UsingRepresentations, Seneviratne2018IdentifyingRecords}. This process involves careful \emph{feature engineering}, whereby features from the \glspl*{ehr} are treated and extracted based on prior knowledge by domain experts. Such methods have managed to achieve accurate predictions on very specific \gls*{dpm} tasks, such as pancreatic cancer prediction \citep{Cherry2020ACancer.}, breast cancer prognosis \citep{Ferroni2019BreastApproach}, cervival cancer prediction \citep{Weegar2020UsingRepresentations}, and metastatic prostate cancer prediction \citep{Seneviratne2018IdentifyingRecords}. However, such methods suffer from issues of generalisability, including the fact that 
 the selection of features by domain experts may prevent discovering new features that could help explain the disease progression. 
 Additionally, ignoring the time dimension
 discards critical information on how the features evolve with time, thus preventing opportunities to learn the temporal patterns in \glspl*{ehr}.

A major challenge in learning from \glspl*{ehr} is thus to incorporate the temporal dimension of the data. Until now, several methods have considered only the \emph{temporal ordering} of the clinical codes, discarding the irregular time intervals between timestamps. Those approaches have enabled the application of a wide array of computational methods, mostly inspired by research in \gls*{nlp}, where \glspl*{ehr} are seen as `text' formed by a succession of `words' (medical codes). 
Inspired by the eponymous \texttt{Doc2Vec} algorithm~\citep{le2014distributed} for \gls*{nlp},
 \citet{Choi2016Multi-layerConcepts} proposed \texttt{Med2Vec}, which learns latent clinical codes and visit-level representations.  In the same year, \citet{Choi2016RETAIN:Mechanism} developed the \texttt{RETAIN} model that employs an attention mechanism integrated with \gls*{rnn} to learn to attend patient visits in reverse time order 
 mimicking the behaviour of physicians when screening patient history. Later on, \citet{Choi2017GRAM:Learning} presented \texttt{GRAM}, a method that employs an attention mechanism that has access to a medical ontology that organises the clinical codes into a hierarchical structure. \texttt{GRAM} is motivated when the dataset contains clinical codes that are sparse and rare, mitigating the risk of model overfitting on those rare codes. In \citep{Choi2018MiME:Healthcare}, \texttt{MIME} is presented as a multilevel representation learning method. The clinical codes in each visit are separated into  treatment codes and diagnosis codes, each fed into different passes to the model.  In a task for heart failure prediction, \texttt{MIME} outperformed both \texttt{Med2Vec} \citep{Choi2016Multi-layerConcepts} and \texttt{GRAM} \citep{Choi2017GRAM:Learning}, which were both trained only on diagnostic codes.

Although several of the aforementioned methods exploit the temporal ordering of visits, none of them use the length of delays between the visits and its potential impact in describing the patient state.  
%
\citet{Zhang2018Patient2Vec:Record} proposed \texttt{Patient2Vec}, which applies time-binning to patient visits and then employs a double attention mechanism: one to produce latent representations from the clinical codes within each time bin, and a second one to integrate the latent representation of the time bins into a sequence using a \gls*{gru}. 
%
%
%
In the task of predicting future hospitalisation from an \gls*{ehr}, \texttt{Patient2Vec} \citep{Zhang2018Patient2Vec:Record} outperformed \texttt{RETAIN} \citep{Choi2016RETAIN:Mechanism}. Although \texttt{Patient2Vec} uses some of the temporal information, time-binning still imposes strong assumptions on the structure of the data.
In addition, \citet{Cai2018MedicalAttention} developed a time-aware attention mechanism and \citet{Peng2019TemporalEmbedding} developed the \texttt{TeSAN} model; these two approaches exploit the time gaps between successive patient visits based on the neural attention mechanism. 

Our method departs from these methods by employing neural \glspl*{ode} to model the temporal evolution of the medical conditions without over-engineering the attention mechanism,  yet explicitly incorporating the time-intervals between consecutive timestamps.
Our proposed method strives to provide a simple framework that (i) flexibly incorporates several information sources, such as timestamped procedures and timestamped numerical lab tests, and (ii) provides a simple, direct reconstruction of time-continuous risk trajectories between the consecutive timestamps.

\section{Methods}

In this section, we first describe the two main constituents of \verb|ICE-NODE|: neural \glspl*{ode} and medical code embeddings. After that, We introduce the \verb|ICE-NODE| model and architecture. 

\paragraph*{Notation.} Throughout, we use $[\bm{A},\bm{B}]$ to denote the horizontal concatenation of two vectors or matrices, and $[\bm{A};\bm{B}]$ for vertical concatenation. When writing functions, we use a semicolon to separate the function input variables from its parameters, e.g., $f(\bm{x}; \bm{\theta})$. 

\subsection{Neural \glspl*{ode}}

We use neural \glspl*{ode} to learn from timestamped diagnostic codes from \glspl*{ehr}.
In contrast to \gls*{rnn} models, which only use the temporal ordering of medical codes, \glspl*{ode} capture the irregular intervals between timestamps, which range from days to months or even years.  

We begin our design by assuming that each patient is described by a time-continuous hidden state $\bm{h}(t) \in \mathbb{R}^{d_h}$ with $d_h$ dimensions. A system of \glspl*{ode} is used to model the temporal evolution of $\bm{h}(t)$ as:
\begin{equation}\label{eq:ode}
    \dv{\bm{h}(t)}{t} = f_d \left(\bm{h}(t); \, \bm{\theta}_d \right)
\end{equation}
where $f_d: \mathbb{R}^{d_h} \mapsto \mathbb{R}^{d_h}$ is the dynamics function parametrised by $\bm{\theta}_d \in \mathbb{R}^{d_p}$, a vector of $d_p$ parameters that can be learnt from data, as shown below.

To evolve the hidden state $\bm{h}(t)$ from timestamp $t_0$ to $t_1$, we write the following \gls*{ivp}:
\begin{equation}\label{eq:ivp}
    \bm{h}(t_1) = \bm{h}(t_0) + \int_{t_0}^{t_1} f_d \left(\bm{h}(t); \, \bm{\theta}_d \right) \, dt \, ,
\end{equation}
where $\bm{h}(t_0)$ is the initial hidden state at $t_0$, to be estimated from the data as shown in Section~\ref{subsec:arch}. 
In general, the \gls*{ivp} in Eq.~\eqref{eq:ivp} is solved numerically at all timestamps $t_k$ using an \verb|IVPSolve| routine:
\begin{equation}
    \bm{h}(t_k) = \text{\texttt{IVPSolve}} 
    \left(f_d, \bm{\theta}_d, \bm{h}(t_{k-1}), [t_{k-1}, t_k] \right) + \varepsilon\, , \label{eq:ivpsolve}
\end{equation}
where $\varepsilon$ is the numerical approximation error. In this work, we use adaptive-step Runge-Kutta 4(5)~\citep{dormand1980family} to perform the numerical integrations.  

 In Section~\ref{subsec:arch}, we develop a model that learns to predict a diagnostic code at time $t$ from the patient state $\bm{h}(t)$, which is also learnt from the data.
The algorithm therefore uses the data (i.e., the observed diagnostic codes collected at timestamps $t_k$) to learn the parameters $\bm{\theta}^*_d$ of the dynamics function in Eq.~\eqref{eq:ode} by minimising a loss function over the whole dataset with terms 
\[ \mathcal{L}(\bm{h}(t_k)) = \mathcal{L}\left( \text{\texttt{IVPSolve}}(f_d, \bm{\theta}_d, \bm{h}(t_{k-1}), [t_{k-1}, t_k]) \right).
\]
The loss function is minimised using \gls*{sgd}, which requires computing the gradient $\nabla_{\bm{\theta}_d} \mathcal{L}$ using deep learning libraries (e.g., \verb|PyTorch| or \verb|JAX|) with reverse-mode \gls*{ad}~\citep{baydin2018automatic}. \gls*{ad} efficiently constructs a computational graph linking the inputs of the function with all the intermediate operations leading to the output of the function, on which backpropagation can then be applied to obtain the gradient.   
However, when applied to a loss function that depends on an \gls*{ode} solver~\eqref{eq:ivpsolve}, this requires storing all the intermediate values of the \verb|IVPSolve| iterations, which may cause problems with computer memory.  
Fortunately, an efficient algorithm~\citep{NEURIPS2018_69386f6b} based on the \emph{adjoint method} computes the gradient
using a time-backward \gls*{ivp} from $t_k$ to $t_{k-1}$
and only requires storing the final value $\bm{h}(t_k)$.

\subsection{Clinical Embeddings module}

Central to our tasks here is to learn highly informative embeddings for the clinical codes that conform our data.  Let us consider the set of all possible clinical codes, $\mathcal{C}$, with cardinality $C=|\mathcal{C}|$. 
The aim of the embedding is to find a transformation that represents a set of clinical codes $c \subseteq  \mathcal{C}$ through a fixed-size vector $\bm{g} \in \mathbb{R}^{d_e}$, which provides a compact representation that retains high information content about the clinical codes.
We start with a multi-hot encoding of the set of clinical codes $c$ as a binary vector $\bm{v} \in \{0, 1\}^C$ with coordinates equal to 1 for the codes present in $c$ and coordinates equal to 0 elsewhere. This large, sparse, binary vector $\bm{v}$ is then transformed into a compact representation $\bm{g}$, i.e., an embedding is obtained. There is a large array of embedding techniques in the literature. We now describe briefly the two methods used in this work.

\paragraph{Simple matrix embedding.} A straightforward approach is to transform $\bm{v} \in \{0, 1\}^C$ into a compact representation $\bm{g} \in \mathbb{R}^{d_e}$ by using an affine map $f_M: \{0, 1\}^C \mapsto \mathbb{R}^{d_e}$:
\begin{equation}\label{eq:mat-emb}
    \bm{g} = f_M(\bm{v}; \bm{\theta}_M) = \bm{W}_M \, \bm{v} + \bm{b}_M
\end{equation}
where $\bm{W}_M \in \mathbb{R}^{d_e \times C}$ is a learnable matrix, $\bm{b}_M \in \mathbb{R}^{d_e}$ is a learnable vector, and $\bm{\theta}_M$ denotes the concatenation $[\bm{W}_M, \bm{b}_M]$.

\paragraph{GRaph-based Attention Model (\texttt{GRAM}) embedding.} 
Several of the most widely used clinical coding schemes, such as \emph{SNOMED-CT} and \emph{ICD}, can be organised into a medical ontology, i.e., a hierarchy of codes where high level (parent) codes represent abstract medical concepts and, as we go down the hierarchy, children codes represent increasingly detailed and precisely described medical concepts. 
Mathematically, this hierarchy is represented by a \gls*{dag}, which we denote as $\mathcal{G}$. 
Choi et. al ~\citep{Choi2017GRAM:Learning} proposed the \verb|GRAM| algorithm, where each medical concept is represented as a convex combination of vectors of `basic embeddings' of the code itself and all of its ancestors in $\mathcal{G}$. This approach enriches the embedding
by retaining information about the ancestry of the code, and mitigates the risk of overfitting rare medical concepts in the training dataset. 
In summary, the method obtains a set of basic embeddings  $\bm{\theta}_E = [\bm{e}_1; \ldots; \bm{e}_C] \in \mathbb{R}^{C \times d_e}$, one for each medical code $c_i \in \mathcal{C}$. 
(To initialise the learning process, the basic embeddings $\bm{\theta}_E$ can be randomly initialised, or obtained with \emph{GloVe}~\citep{pennington2014glove}.) 
The final embedding of a medical code $c_i$ is given by the convex combination:
\begin{equation}\label{eq:convex}
    \bm{g}_i = \sum_{j \in  \mathcal{A}(i)} \alpha_{ij} \bm{e}_j, \quad \text{with} \quad  \sum_{j \in  \mathcal{A}(i)} \alpha_{ij} = 1, \quad \alpha_{ij} \geq 0, 
\end{equation}
where $\mathcal{A}(i)$ is the set of coordinates for the union of the medical concept $c_i$ and all of its ancestors in $\mathcal{G}$. The weights $\alpha_{ij} \in \mathbb{R}$ are computed with the \textit{softmax function}:
\begin{equation}
    \alpha_{ij} = \frac{\exp\left(f_R(\bm{e}_i, \bm{e}_j)\right)}{\sum_{k \in  \mathcal{A}(i)} \exp \left(f_R (\bm{e}_i, \bm{e}_k)\right)},
\end{equation} 
where the \emph{self-attention} $f_R: \mathbb{R}^{d_e} \times \mathbb{R}^{d_e} \mapsto \mathbb{R}$ estimates the relatedness between the embeddings. Originally, the authors implemented the function $f_R$ as a \gls*{mlp} with a single hidden layer of size $\ell$:
\begin{equation}\label{eq:f_tanh}
    f_R(\bm{e}_i, \bm{e}_j; \bm{\theta}_R) = \bm{u}_R^T\tanh(\bm{W}_R  \begin{bmatrix}
           \bm{e}_i \\
           \bm{e}_j
         \end{bmatrix} + \bm{b}_R),
\end{equation}
where $\bm{W}_R \in \mathbb{R}^{\ell \times 2d_e}$, $\bm{b}_R \in \mathbb{R}^\ell$ 
and $\bm{u}_R \in \mathbb{R}^\ell$. 
Here, $\bm{\theta}_R$ denotes $[\bm{W}_R, \bm{b}_R, \bm{u}_R]$. 
Note that the vertical concatenation $[\bm{e}_i;\bm{e}_j]$ preserves the child-ancestor order. 
In this work, we also used a less parametrised variant suggested in ~\cite[Appendix F.2]{pmlr-v139-kim21i} with stability guarantees:
\begin{equation}\label{eq:f_l2}
    f_r(\bm{e}_i, \bm{e}_j; \bm{\theta}_r) = \exp\left(-\frac{|| (\bm{e}_i - \bm{e}_j)^T \bm{\theta}_r||^2_2}{ \sqrt{\ell}} \right),
\end{equation}
where $\bm{\theta}_r \in \mathbb{R}^{\ell \times d_e}$.

Given basic embeddings $\bm{\theta}_E$ and attention parameters $\bm{\theta}_R$ (or $\bm{\theta}_r$ if \eqref{eq:f_l2} is used), the embedding of the codes $c_i \in \mathcal{C}$ is given by the matrix 
\[\bm{G}_{(\bm{\theta_E}, \bm{\theta_R})} = [\bm{g}_1; \ldots; \bm{g}_C] \in \mathbb{R}^{C \times d_e}.
\]
To transform the multi-hot binary vector $\bm{v} \in \{0, 1\}^C$, representing the set of multiple medical concepts $c \subseteq \mathcal{C}$, into the embedding space we just apply the transformation
\begin{equation}\label{eq:gram}
    \bm{g} = f_G(\bm{v}; \bm{\theta_E}, \bm{\theta}_R) = \tanh{\big(\bm{v} \, \bm{G}_{(\bm{\theta_E}, \bm{\theta_R})}\big)}
\end{equation}

\paragraph{Remark.} \verb|ICE-NODE| does not assume a particular embedding method; hence $\bm{g}$ (i.e., the embedding of $\bm{v}$) can be computed via~\eqref{eq:mat-emb} or~\eqref{eq:gram}. The particular choice in the numerical experiments is made explicit in \sectionref{sec:experiments}.

\subsection{The \texttt{ICE-NODE} model}\label{subsec:arch}


\verb|ICE-NODE| is the architecture that we propose to model the timestamped clinical codes contained in \glspl*{ehr} using neural \glspl*{ode} coupled with a clinical code embedding module to predict diagnostic codes at a given future timestamp $t_f$. 

\paragraph*{Notation.}
Each patient $i$ is represented as a sequence of timestamped clinical codes $\{\big(t_k, c(i, t_k)\big)\}_{t_k \in \mathcal{T}(i)}$, where $t_k$ is the $k$-th timestamp in the temporally-ordered set of timestamps $\mathcal{T}(i)$ in the \gls*{ehr} of patient $i$, and $c(i, t_k) \subseteq \mathcal{C}$ is the set of medical concepts present at $t_k$ for this particular patient. To simplify notation, we will drop the patient index $i$ when suitable. The binary (multi-hot) representation of $\{\big(t_k, c(t_k)\big)\}_{t_k \in \mathcal{T}}$ is denoted as  $\{\big(t_k, \bm{v}(t_k)\big)\}_{t_k \in \mathcal{T}}$. 
The embedding module transforms each set of clinical codes $c(t_k)$ into their embeddings $\bm{g}(t_k)$, and hence we obtain the set of timestamped embeddings $\{\big(t_k, \bm{g}(t_k)\big)\}_{t_k \in \mathcal{T}}$. The $d_h$-dimensional hidden state for the patient $\bm{h}(t)=[\bm{h}_m(t); \bm{h}_e(t)]$ is structured to consist of two components: a memory state $\bm{h}_m(t) \in \mathbb{R}^{d_m}$ and an embedding state $\bm{h}_e(t) \in \mathbb{R}^{d_e}$, such that $d_m + d_e = d_h$.

\begin{figure}[t]
    \centering
    \includegraphics[width=.95\linewidth]{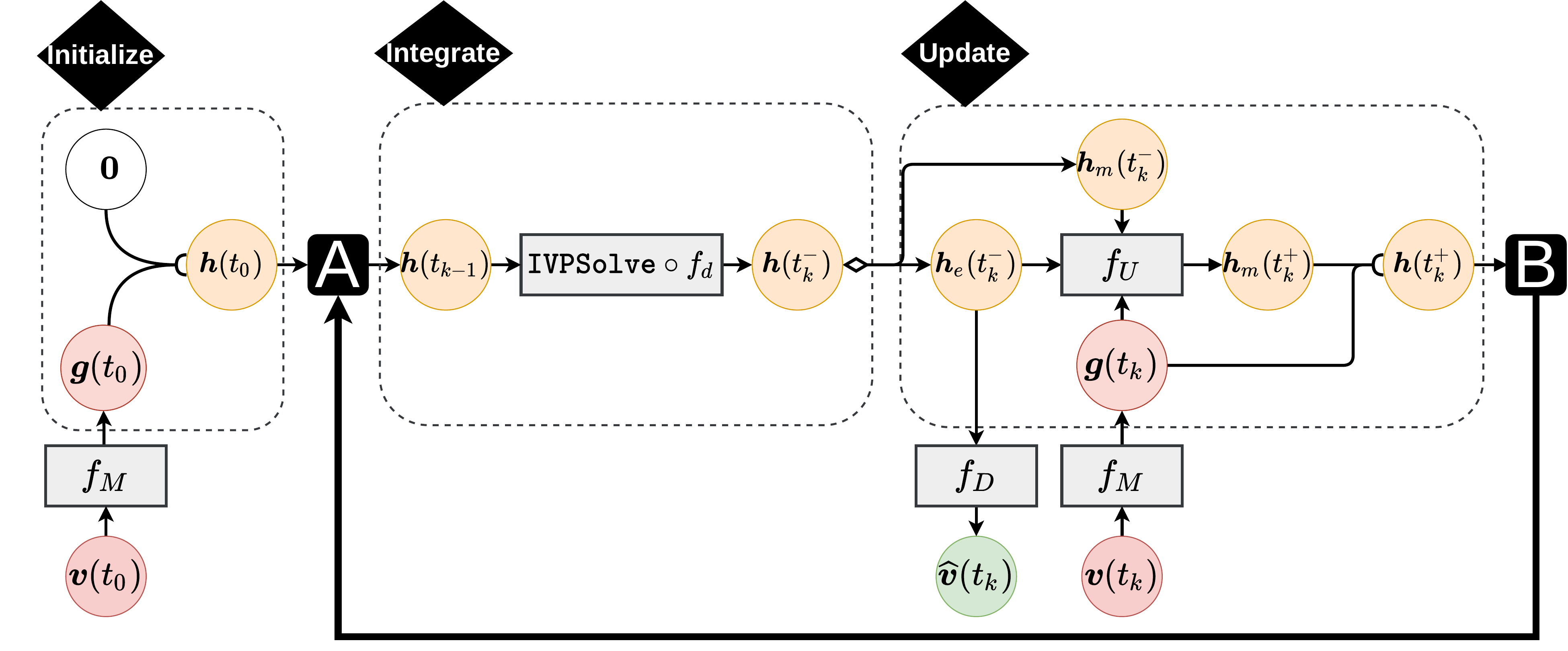}
       \caption{\footnotesize \textbf{Schematic representation of the \texttt{ICE-NODE} framework to model patient histories of timestamped clinical concepts, as recorded in their \gls*{ehr}.} 
    The health state $\bm{h}(t_{k-1})$ is \emph{integrated} with an \gls*{ode} solver and learnable dynamics function $f_d$ to obtain a future version $\bm{h}(t_k^-)$, i.e. just before time $t_k$. 
    The memory state $\bm{h}_m(t_k^-)$ is then \emph{updated} to accommodate the new information at time $t_k$ by applying an update transformation $f_U$, which incorporates the memory state itself $\bm{h}_m(t_k^-)$, the integrated embedding state $\bm{h}_e(t_k^-)$, and the embeddings of the clinical concepts $\bm{g}(t_k)$ at the timestamp $t_k$. This `Integrate-Update' cycle is repeated for subsequent timestamps.} 
    \label{fig:icenode}
\end{figure}

\subsubsection{\texttt{ICE-NODE} Architecture}

The initial hidden state $\bm{h}(t_0) \in \mathbb{R}^{d_h}$ at the first timestamp $t_0$ is given by $\bm{h}(t_0) = [\bm{h}_m(t_0); \bm{h}_e(t_0) ] =[\bm{0}; \bm{g}(t_0)]$, where the memory state is initialised with zeros and the embedding state is initialised with the embedding of the set of medical codes observed at $t_0$. 

We next evolve $\bm{h}(t_0)$ by integration over $[t_0, t_1)$, i.e., to a time $t_1^-$ just before $t_1$:
\begin{equation}\label{eq:odeint}
\bm{h}(t_1^-) = \begin{bmatrix}
           \bm{h}_m(t_1^-) \\
            \bm{h}_e(t_1^-)
         \end{bmatrix} = \text{\texttt{IVPSolve}}(f_d, \bm{\theta}_d, \bm{h}(t_0), [t_0, t_1]).
\end{equation}
To this evolved state, we apply a decoding function $f_{D}: \mathbb{R}^{d_e} \mapsto [0, 1]^C$ to obtain the predicted  clinical codes at time $t_1$:
\begin{equation}\label{eq:dec}
   \widehat{\bm{v}}(t_1) = f_D(\bm{h}_e(t_1^-); \bm{\theta}_D).
\end{equation}
This prediction can then be scored against the observed codes $\bm{v}(t_1)$ in the dataset. 

To accommodate the new information gathered at each timestamp into  $\bm{h}(t)$, we update the memory state:
\begin{equation}\label{eq:update}
\bm{h}_m(t_1^+) = f_U(\bm{h}_m(t_1^-), \bm{h}_e(t_1^-), \bm{g}(t_1); \bm{\theta}_U),
\end{equation}
where $t_1^+$ is a time just after the observation timestamp $t_1$, and $f_U: \mathbb{R}^{d_m} \times \mathbb{R}^{d_e} \times \mathbb{R}^{d_e} \mapsto \mathbb{R}^{d_m}$ adjusts the memory state $\bm{h}_m(t_1^-)$ to incorporate the new information at $t_1$.
The steps \eqref{eq:odeint}-\eqref{eq:dec}-\eqref{eq:update} are then repeated for all remaining timestamps in $\mathcal{T}$. 

After fitting all the timestamps in the history of a patient, we apply a final integration~\eqref{eq:odeint} from the last timestamp in $\mathcal{T}$ until the future timestamp $t_f$, to obtain the hidden health state $\bm{h}(t_f)$. This is followed by the decoding step \eqref{eq:dec} to produce the diagnostic predictions $\widehat{\bm{v}}(t_f)$.

\subsubsection{Training \texttt{ICE-NODE}}

%
%
The training is carried out by minimising the following loss function, which includes the cross-entropy comparing the predictions $\widehat{\bm{v}}(t_k)$ and the observations $\bm{v}(t_k)$ of the codes at timestamps  $t_k \in \{t_1, \ldots, t_{|\mathcal{T}|-1}\}$ and a regularisation term that enforces the smoothness of the dynamics: 
\begin{equation}\label{eq:icenode_loss}
    \mathcal{L} = \frac{1}{|\mathcal{T}|-1}  \sum_{k=1}^{|\mathcal{T}-1|} \mathcal{L}_{\bm{v}}(t_k) + \alpha_K \mathcal{R}_K(t_k; \bm{\theta}_d),
\end{equation}
where the cross-entropies are:
\begin{equation}\label{eq:pred_loss}
    \mathcal{L}_{\bm{v}}(t_k) = \bm{v}(t_k)^T \log{(\widehat{\bm{v}}(t_k))} + (\bm{1}-\bm{v}(t_k))^T\log{(\bm{1} - \widehat{\bm{v}}(t_k))}, 
\end{equation}
and the regularisation term is given by:
\begin{equation}\label{eq:dyn_reg}
    \mathcal{R}_K(t_k; \bm{\theta}_d) = \int_{t_{k-1}}^{t_k} 
    \left \| \dv[K]{\bm{h}(t)}{t} \right \| ^2_2\,dt \, .
\end{equation}
The regularisation term $\mathcal{R}_K$~\citep{Kelly2020LearningSolve}, which is inspired by insights from adaptive-step solvers, 
diminishes the magnitude of the higher order derivatives and increases both trajectory smoothness and step size. This allows us to efficiently concentrate on a smaller subspace of dynamic trajectories. 
In general, the hyperparameter $K$ must be no larger than $m$, the order of the \gls*{ode} solver (Runge-Kutta with $m=4$ here), but since we further assume that the health state $\bm{h}(t)$ has vanishing derivatives above second-order, we set $K=3$.
Finally, $\alpha_K$ is the hyperparameter that sets the relative weight of the regularisation term~\eqref{eq:dyn_reg}. It is set here to a large value of $\alpha_K=1000$ to enforce vanishing derivatives of order $K\leq 3$.

For each training iteration, we randomly sample (with replacement) a fixed number of patients $B$ from the total of $N$ patients in the training set. The loss~\eqref{eq:icenode_loss} is then averaged over the $B$ sampled patients and the gradients are computed, which are then passed to the optimiser to update the parameters.
Here, we use the \emph{Adam} optimiser~\citep{kingma2014adam}, which achieves remarkably better convergence than both \gls*{sgd} and \emph{Adamax}~\citep{kingma2014adam}. 
We find that convergence is improved by using two independent learning rates for the \emph{Adam} optimiser: one for the dynamics parameters, and a separate one for the other parameters of \verb|ICE-NODE|. \appendixref{appendix:config} describes in detail the training settings.
We also describe in \appendixref{appendix:config} the strategy for hyperparameter optimisation, including the use of the \verb|optuna| framework \citep{optuna_2019}, to settle on an optimal choice of functions and their configuration for the dynamics function $f_d$~\eqref{eq:odeint}, the decoding function $f_D$~\eqref{eq:dec}, and the update function $f_U$~\eqref{eq:update}. 


\section{Experiments}\label{sec:experiments}

\subsection{Description of the datasets}

\paragraph{\acrshort*{mimic3}~\citep{mimic3, johnson2016mimic}:}
\gls*{mimic3} is a publicly available dataset that contains \glspl*{ehr} for over 46.5K patients who had at least one admission at \gls*{bidmc} between 2001 and 2012. For each admission, the patient can be discharged on the same day or stay for a longer time,
but in all cases \gls*{mimic3} stores all the diagnosis codes related to the entire stay and links them to the discharge timestamp.
This manner of information collection adds a degree of uncertainty to the timestamps associated with long stays; if a patient stays for two months, they will be discharged with a set of diagnosis codes, but this does not indicate precisely when the diagnoses were made during those two months. 
We therefore restrict our study to patients (i) who had at least two admissions, and (ii) whose admissions were all at most 2 weeks long.
With these restrictions, our dataset includes 4.4K patients. \gls*{mimic3} uses the \emph{ICD-9} coding scheme to store the diagnosis codes, which we convert into the \gls*{ccs}\footnote{\url{https://www.hcup-us.ahrq.gov/toolssoftware/ccs/ccs.jsp}} multi-level (i.e. hierarchical) scheme. This conversion reduces substantially the complexity of the data---whereas \emph{ICD-9} provides over 15K diagnosis codes, \gls*{ccs} contains 589 diagnosis codes.
When we use \verb|GRAM| embeddings, we thus employ the corresponding \gls*{dag} $\mathcal{G}$ of the \gls*{ccs} hierarchical scheme for the analysis of the set of clinical codes $\mathcal{C}$. 

\paragraph{\acrshort*{mimic4}~\citep{mimic4}:}
\gls*{mimic4} is another publicly available dataset that contains \glspl*{ehr} for over 256K patients who had a critical care or emergency admission in \gls*{bidmc} between 2008 and 2019. 
After applying the same rules for patient selection as for \gls*{mimic3}, we end up having \glspl*{ehr} for over 70K patients. 
\gls*{mimic4} uses the \emph{ICD-10} coding scheme, which we map to \emph{ICD-9}\footnote{\url{https://www.nber.org/research/data/icd-9-cm-and-icd-10-cm-and-icd-10-pcs-crosswalk-or-general-equivalence-mappings}} and eventually to the \gls*{ccs} coding scheme, so that models trained on \gls*{mimic3} can be tested on \gls*{mimic4} and vice versa. 

\paragraph{Summary:}
Table~\ref{tab:data_stats} presents descriptive statistics of the two datasets constructed from \gls*{mimic3} and \gls*{mimic4}. For both datasets, we randomly split patients into training:validation:testing with ratios $0.70:0.15:0.15$. \appendixref{appendix:consort} includes consort diagrams for the extraction of the training-validation-test splits from both datasets.

\begin{table}[t]
    \centering
    \caption{\footnotesize Summary statistics for the two datasets used in our experiments.}
    \begin{tabular}{l|ll}
        \toprule
         Statistics &  \gls*{mimic3} & \gls*{mimic4} \\
         \midrule 
         No.\ of patients ($N_p$) &  4,385  &  70,027 \\
         No.\ of admissions ($N_a$) & 10,954 & 265,637 \\
         Avg. admissions per patient ($N_a/N_p$) & $2.47$ &  $3.66$ \\
         Avg.\ ($\pm$std.) weeks between admissions & $66.1 \, (\pm97.3)$ & $53.8 \, (\pm85.8)$ \\ 
         Avg.\ ($\pm$std.) days of stay & $5.9 \, (\pm3.5)$ & $3.2 \, (\pm2.8)$\\
         Avg.\ no.\ of ICD-9 diagnostic discharge codes & $11.65$ & $11.46$ \\
         Avg.\ no.\ of \gls*{ccs} diagnostic discharge codes & $10.84$ & $9.75$ \\
        \bottomrule
    \end{tabular}
    \label{tab:data_stats}
\end{table}

\subsection{Methods for benchmarking}


Our numerical experiments have been compared and benchmarked against state-of-the-art baseline methods. In most cases, we have implemented a version that uses matrix embeddings~\eqref{eq:mat-emb} and a version that uses \verb|GRAM| embeddings~\eqref{eq:gram}. The latter versions are denoted by adding the letter \verb|/G| to the corresponding acronyms. \\

\noindent The baseline methods considered are:

\paragraph{\texttt{LogReg}:}
We implement a standard logistic regression with elastic net regularisation. This method takes as an input a binary vector encoding the occurrence of diagnostic codes in the past. We consider this model as a representative of models that do not exploit the sequential nor the full temporal information of the \glspl*{ehr}.

\paragraph{\texttt{RETAIN}:} 
This method~\citep{Choi2016RETAIN:Mechanism} learns patient visits in reverse time order using an attention mechanism using the temporal ordering of the codes, but ignoring the irregular time intervals between them. 

\paragraph{\texttt{\acrshort*{gru}} \& \texttt{\acrshort*{gru}/G}:}
Based on the architecture developed by~\citet{Choi2017GRAM:Learning}, we have implemented two versions: the original one with \verb|GRAM| embeddings, denoted \texttt{\acrshort*{gru}/G}, and another one with matrix embeddings, denoted \texttt{\acrshort*{gru}}. Again, this method only uses the time ordered information but not the full temporal information in the data. 
\\

\noindent These baseline methods are compared against our model in different versions:
\paragraph{\texttt{ICE-NODE} \& \texttt{ICE-NODE/G}: }
We use our model, as developed above, with both matrix embeddings and \texttt{GRAM} embeddings, using the full temporal information available in the \glspl*{ehr}.

\paragraph{\texttt{ICE-NODE\_UNIFORM} \& \texttt{ICE-NODE\_UNIFORM/G}: } We have also considered a variant of \texttt{ICE-NODE} where we fix the intervals between consecutive timestamps to be one week in two versions: one with matrix embeddings, denoted \texttt{ICE-NODE\_UNIFORM}, and one with \texttt{GRAM} embeddings, denoted \texttt{ICE-NODE\_UNIFORM/G}.  
These versions ignore the irregularity of the temporal sampling, and just preserve the ordering.
We use this variant particularly to assess whether a clinical code prediction is improved by incorporating the full information of the timestamps, or if using only the temporal ordering is sufficient.

\subsection{Analysis of prediction performance}

We analysed the different versions of our proposed model along with the baseline models through three experiments: 
\begin{itemize}
    \item \textbf{Experiment A}: Training and testing on \gls*{mimic3}
    \item \textbf{Experiment B}: Training and testing on \gls*{mimic4}
    \item \textbf{Experiment C}: Training on \gls*{mimic4} followed by testing on the entire \gls*{mimic3}
\end{itemize}

Evaluating the prediction performance of disease progression models with respect to a large number of clinical codes is a nontrivial problem. While one model can be the most competent in predicting a particular set of clinical codes, the same model can be outperformed by another model in predicting a different set of clinical codes. Typically, researchers often focus on a specific category of diseases, and compare the predictability between multiple models. 

Alternatively, other evaluations are geared towards systematically evaluating all medical codes, yet controlling for their different frequency, since predicting very common medical codes can be less informative.
%
%
 Clinical codes are then partitioned into quantiles according to their frequency in the dataset, such that the model predictability can be estimated for different percentile ranges separately, from the infrequent to the most frequent codes. With this evaluation method, \citet{Choi2017GRAM:Learning} showed that their algorithm (denoted here as \texttt{GRU/G}) is specifically competent in predicting clinical codes that are observed rarely in the training data. 
 We have applied this approach to evaluate the performance of the methods for Experiments A and B. We report the results in Table~\ref{tab:top15acc}, where we show the averaged top-15 prediction accuracy (i.e., for each clinical code, we score 1 if it is correctly detected in the top-15 predictions by the model at each visit, and 0 otherwise).
We find that \texttt{ICE-NODE} is particularly competent in predicting codes that are infrequent in the training dataset (i.e., in the 0-20 and 20-40 quantiles) while still performing well in more frequent codes. 

\begin{table}[htb!]
\centering
\floatconts
    {tab:top15acc}
    {\caption{\footnotesize \textbf{Top-15 prediction accuracy for clinical codes according to their frequency in the training set.} The clinical codes are partitioned into five quantiles, according to their frequency in the training set. 
    For each admission, we score the top-15 codes. The top performing method for each quantile group is highlighted in dark green and the second best method in light green.}}%
    {
    \subtable[\footnotesize Accuracy of Experiment A (training and testing on \gls*{mimic3}).][b]{
      \label{tab:mimic3_acc10}
        \begin{tabular}{l|rrrrr}
        \hline
         \multirow{2}{4em}{Model} & \multicolumn{5}{c}{Clinical codes frequency (quantile ranges)} \\
         \cline{2-6}
         & {0-20} & {20-40} & {40-60} & {60-80} & {80-100} \\
        \hline
        \texttt{LogReg} & 
        0.164 & 
        {\cellcolor[HTML]{276419}} \color[HTML]{F1F1F1} 0.454 &  
        0.544 & 
        0.683 & 
        0.777 \\
        \texttt{RETAIN} & 
        0.208 & 
        0.415 & 
        {\cellcolor[HTML]{276419}} \color[HTML]{F1F1F1} 0.579 & 
        0.778 & 
        {\cellcolor[HTML]{276419}} \color[HTML]{F1F1F1} 0.926 \\
        \texttt{GRU} & 
        {\cellcolor[HTML]{9ACD61}} \color[HTML]{000000} 0.214 & 
        0.400 & 
        0.556 & 
        0.768 & 
        0.900 \\
        \texttt{GRU/G} & 
        0.204 & 
        0.403 & 
        0.546 & 
        0.778 & 
        0.906 \\
        \cdashline{1-6}
        \texttt{ICE-NODE\_UNIFORM} &   0.211 &   0.422 &   0.566 & {\cellcolor[HTML]{276419}} \color[HTML]{F1F1F1} 0.786 & {\cellcolor[HTML]{9ACD61}} \color[HTML]{000000} 0.909 \\
        \texttt{ICE-NODE\_UNIFORM/G} &   0.206 &   0.419 &   0.565 & {\cellcolor[HTML]{9ACD61}} \color[HTML]{000000} 0.782 &   0.907 \\
        \cdashline{1-6}
        \texttt{ICE-NODE} & 
        {\cellcolor[HTML]{276419}} \color[HTML]{F1F1F1} 0.219 & 
        {\cellcolor[HTML]{9ACD61}} \color[HTML]{000000} 0.425 & 
        {\cellcolor[HTML]{9ACD61}} \color[HTML]{000000} 0.578 & 
        {\cellcolor[HTML]{9ACD61}} \color[HTML]{000000} 0.782 & 
        0.902 \\
        \texttt{ICE-NODE/G} & 
        0.205 & 
        0.421 & 
        0.569 & 
        {\cellcolor[HTML]{9ACD61}} \color[HTML]{000000} 0.782 & 
        0.907 \\
        \hline
        \end{tabular} 
    }
    \bigskip
    \vspace*{0.1cm}
    \subtable[\footnotesize Accuracy of Experiment B (training and testing on \gls*{mimic4}).][t]{
        \label{tab:mimic4_acc10}
        \begin{tabular}{l|rrrrr}
        \hline
         \multirow{2}{4em}{Model} & \multicolumn{5}{c}{Clinical codes frequency (quantile ranges)} \\
         \cline{2-6}
         & {0-20} & {20-40} & {40-60} & {60-80} & {80-100} \\
        \hline
        \texttt{LogReg} & 
        0.019 & 
        0.347 & 
        0.606 & 
        {\cellcolor[HTML]{276419}} \color[HTML]{F1F1F1} 0.869 & 
        {\cellcolor[HTML]{276419}} \color[HTML]{F1F1F1} 0.943 \\
        \texttt{RETAIN} & 
        0.362 & 
        0.483 & 
        0.621 & 
        0.806 & 
        {\cellcolor[HTML]{9ACD61}} \color[HTML]{000000} 0.928 \\
        \texttt{GRU} & 
        0.355 & 
        {\cellcolor[HTML]{9ACD61}} \color[HTML]{000000} 0.491 & 
        {\cellcolor[HTML]{276419}} \color[HTML]{F1F1F1} 0.633 & 
        0.808 & 
        0.917 \\
        \texttt{GRU/G} & 
        0.357 & 
        {\cellcolor[HTML]{9ACD61}} \color[HTML]{000000} 0.491 & 
        {\cellcolor[HTML]{9ACD61}} \color[HTML]{000000} 0.628 & 
        {\cellcolor[HTML]{9ACD61}} \color[HTML]{000000} 0.809 & 
        0.921 \\
        \cdashline{1-6}
        \texttt{ICE-NODE\_UNIFORM} & {\cellcolor[HTML]{276419}} \color[HTML]{F1F1F1} 0.376 & 
        0.486 & 
        0.598 & 
        0.773 & 
        0.904 \\
        \texttt{ICE-NODE\_UNIFORM/G} &   
        0.370 & 
        0.485 & 
        0.593 & 
        0.779 & 
        0.903 \\
        \cdashline{1-6}
        \texttt{ICE-NODE} &  
        {\cellcolor[HTML]{9ACD61}} \color[HTML]{000000}
        0.375 & 
        {\cellcolor[HTML]{276419}} \color[HTML]{F1F1F1} 0.493 & 
        0.605 & 
        0.774 & 
        0.904 \\
        \texttt{ICE-NODE/G} & 
        0.374 & 
        {\cellcolor[HTML]{9ACD61}} \color[HTML]{000000} 0.491 & 
        0.601 & 
        0.776 & 
        0.906 \\
        \hline
        \end{tabular}
    }
    }
\end{table}



As an additional measure of performance, we have also carried out a quantification of the \emph{relative competency} at the code level using the \gls*{auc} values achieved by the different methods and the DeLong  test~\citep{delong1988comparing}. This test allows us to establish the statistical significance of the difference between the \gls*{auc} values obtained for any given pair of models. Figure~\ref{fig:relative_competency} summarises the results of this analysis for Experiments A-C.  We find that most well-predicted codes are predicted well by all methods, yet different methods are more competent at predicting particular codes.

\begin{figure}[htb!]
\floatconts
{fig:relative_competency}
{\caption{\footnotesize \textbf{(a)-(c) Relative prediction of medical codes for the different models for Experiments A-C.} \gls*{auc} values of all clinical codes are computed for each model and a DeLong test (\emph{p-value}=0.01) is carried out to establish the statistical significance of the difference in prediction between each pair of models. Clinical codes predicted with \gls*{auc}$>$0.9 by at least one model are assigned to the model with maximum \gls*{auc} and to any other model with no significant difference (according to DeLong test).  Frequencies of clinical codes in the training set are shown at the top. (a) In Experiment A, 27 codes are predicted with an \gls*{auc}$>$0.9 by at least one model: 21 of those codes are predicted equivalently well by all 8 models, whereas the rest are predicted differently by some models. (b) In Experiment B, 84 codes are predicted with an \gls*{auc}$>$0.9 by at least one model; of those, only 4 codes are predicted well by \texttt{LogReg}, whereas 39 codes are predicted equivalently well by the four models that incorporate temporal or sequence information. Some codes are predicted well by several models (39/2/11/2/18); other codes are predicted well only by one model (4/1/3).   (c) In Experiment C, 50 codes are predicted with an \gls*{auc}$>$0.9 by at least one model: 32 codes are predicted equivalently well by all models (but not by \texttt{LogReg}).
There are 4 codes in (b) (`\textit{Pulm hart dx}', `\textit{Other ear dx}', `\textit{Early labor}', `\textit{Forceps del}'), and 1 code in (c) (`\textit{Ac renal fail}') that are only predicted well by \texttt{ICE-NODE} (marked in red).
The \gls*{auc} values for the codes in red are shown in Fig.~\ref{fig:auc_vals} (\appendixref{sec:auc_vals}).
}} %
{%
\subfigure[\footnotesize Experiment A][b]{
        \label{fig:auc_m3}
        \includegraphics[width=0.32\textwidth]{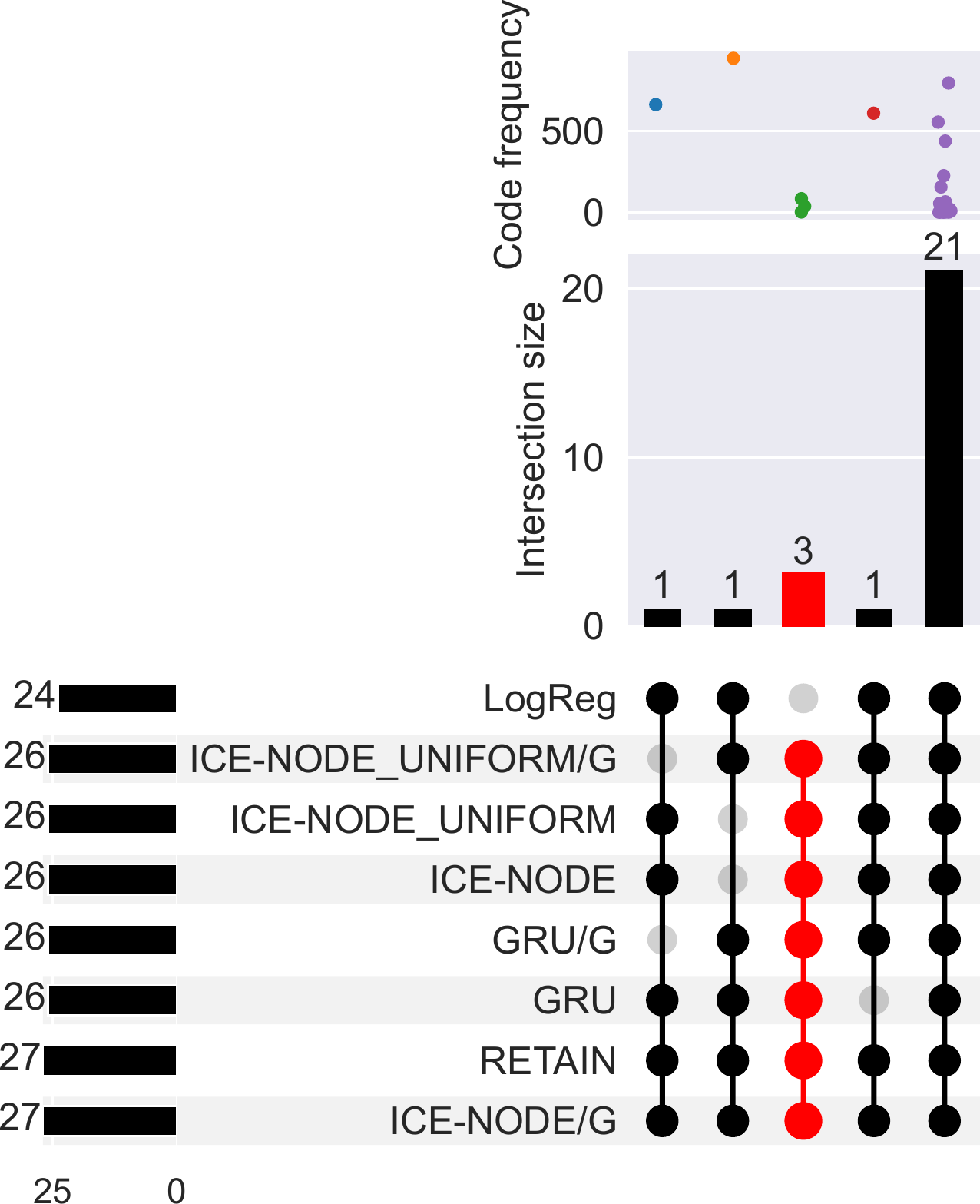}%
} %
\subfigure[\footnotesize Experiment B][b]{%
 \label{fig:auc_m4}
 \includegraphics[width=0.38\textwidth]{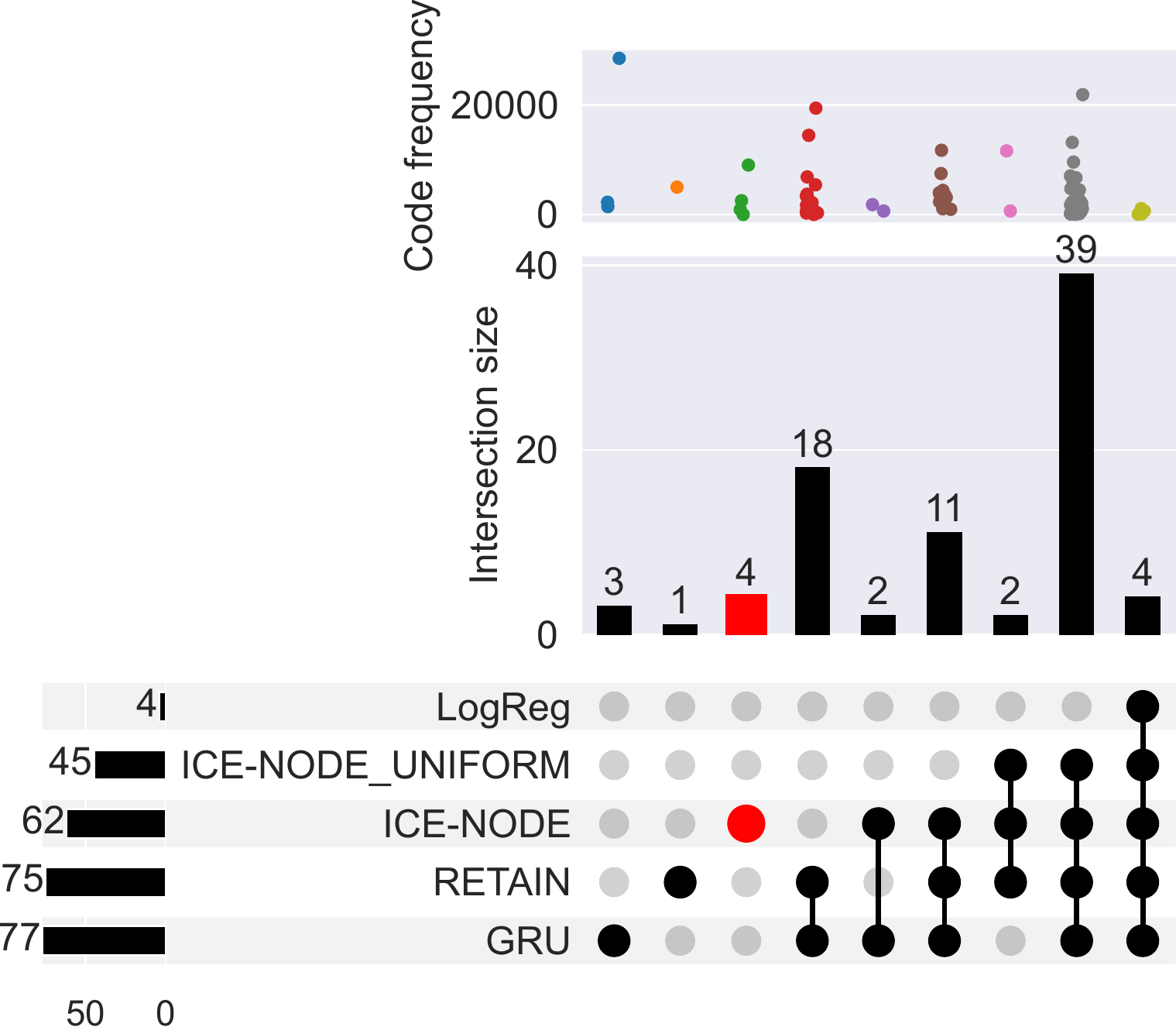}
} %
\subfigure[\footnotesize Experiment C][b]{%
 \label{fig:auc_m4m3}
 \centering{
 \includegraphics[width=0.34\textwidth]{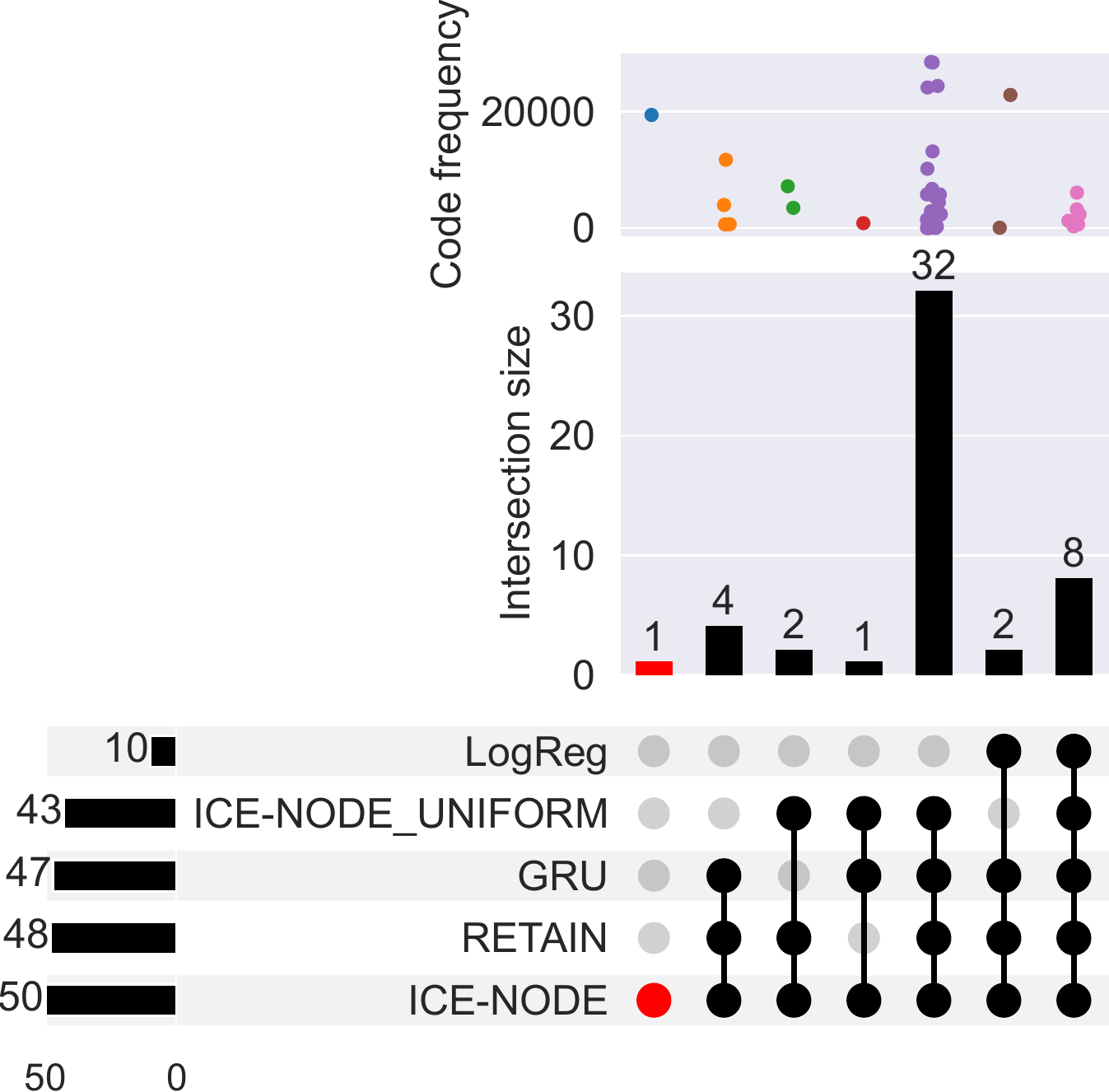}
 }
}%
}
\end{figure}


\subsection{Time-continuous risk trajectories}
Through its use of neural \glspl*{ode}, \texttt{ICE-NODE} stands out from the baseline methods by a qualitative advantage; namely, the possibility of using the learnt hidden patient state $\bm{h}(t)$ to carry out time-continuous predictions of, e.g., disease risk. 
By sampling $\bm{h}(t)$~\eqref{eq:odeint} at arbitrary times, we can apply the prediction function~\eqref{eq:dec} to generate a risk trajectory for a selected patient history. Two examples of such trajectories for two patients in the datasets with conditions where \texttt{ICE-NODE} outperforms the other methods are shown in \figureref{fig:traj_m3_1}, and further examples can be seen in \appendixref{appendix:trajectories}.

\begin{figure}[htb!]
\floatconts
{fig:trajectories}
{\caption{\footnotesize Two predicted risk trajectories for patients in the test set: (a) Predicted risk trajectory of code `\emph{Ac renal fail}' for patient with \texttt{subject\_id=3600} in the test set of \gls*{mimic3}. The history of this patient consists of three hospital stays (i.e., admissions-discharges) with diagnosis of `\emph{Ac renal fail}' at the third discharge. After the first discharge, \texttt{ICE-NODE} predicts a risk for this diagnosis with a probability slightly higher than 0.5. After the second discharge, the risk has jumped to above 0.55 with continuously increasing risk throughout roughly 80 days, before the last hospital stay; (b)  Predicted risk trajectory of `\emph{Pulm hart dx}' for the patient with \texttt{subject\_id=13286711} in the test set of \gls*{mimic4}. In this case, the history consists of five hospital stays with diagnosis of `\emph{Pulm hart dx}' at the fifth and last hospital stay. The risk for this condition kept increasing since the first discharge despite negative reporting for this code throughout.
}}
{
\subfigure[][b]{%
    \label{fig:traj_m3_1}
    \includegraphics[width=.42\linewidth]{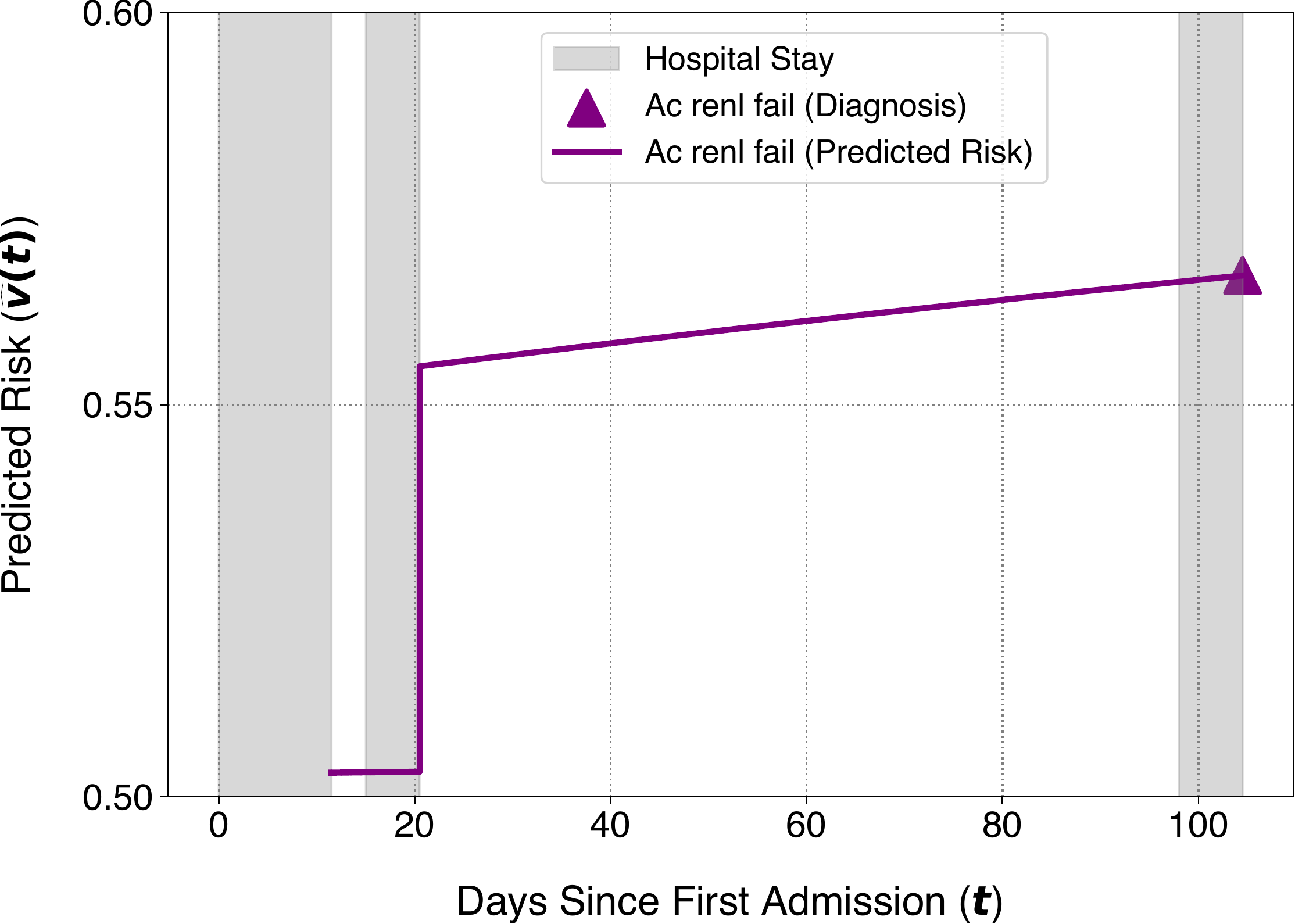}}
{\subfigure[][b]{%
    \label{fig:traj_m4_1}
    \includegraphics[width=.42\linewidth]{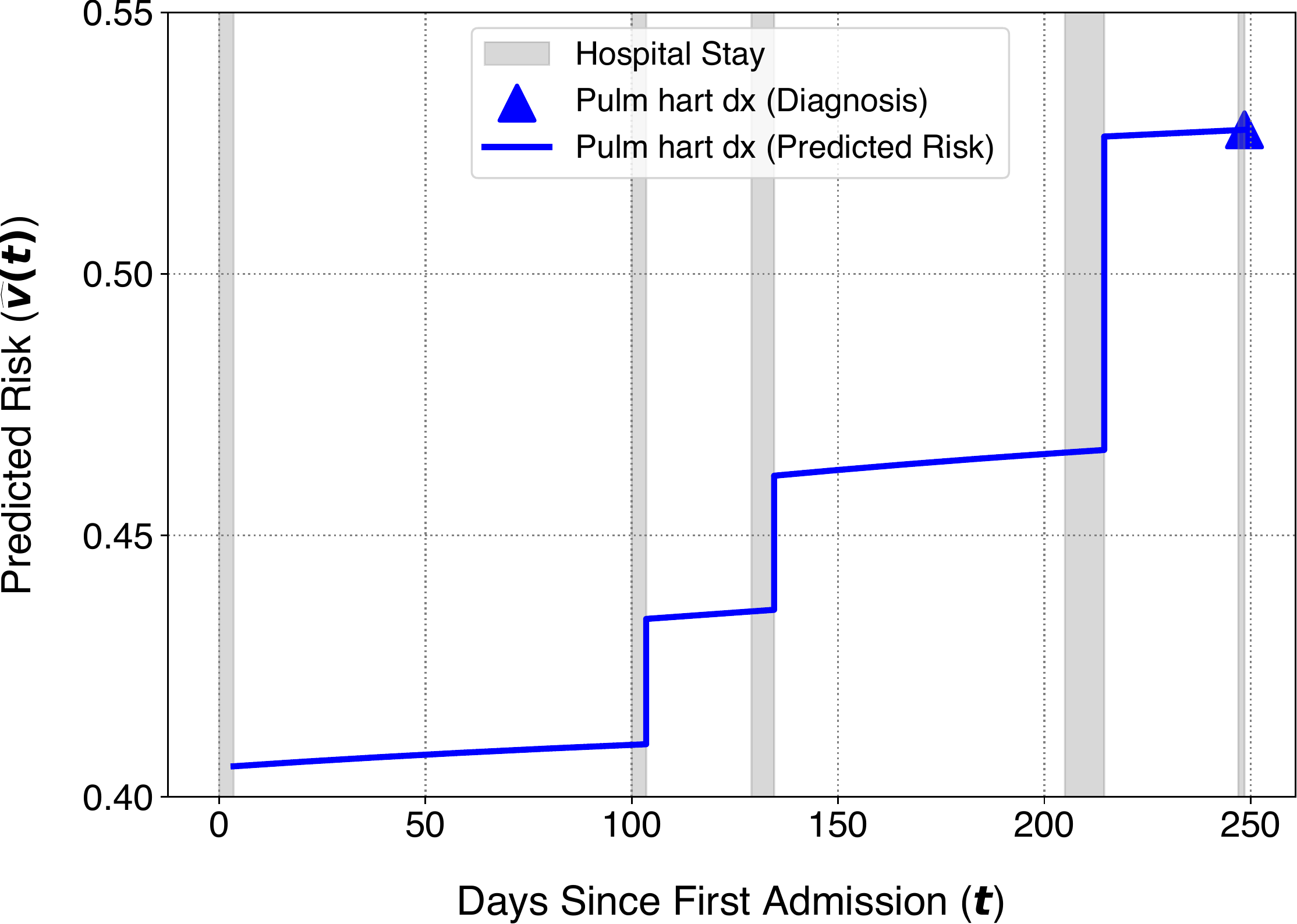}
    }
}
}
\end{figure}

\section{Discussion} 


The results in Table~\ref{tab:top15acc} suggest significant competency of the different variants of \texttt{ICE-NODE}, especially in predicting clinical codes that are rare in the training data. We also find that using \texttt{GRAM} embeddings for \texttt{ICE-NODE} does not achieve the hoped-for improvement over matrix embeddings in our current analysis. However, this initial conclusion might be due to the use of the simple  \gls*{ccs} hierarchical coding scheme with its computationally manageable vocabulary (589 diagnosis codes). However, the use of such a simple coding scheme might have resulted in a loss of information. Therefore, \texttt{ICE-NODE/G} might still add value if more complex hierarchical coding schemes, such as \emph{ICD-9} (15K diagnosis codes), \emph{ICD-10} (70K diagnosis codes), or \emph{SNOMED-CT} (383K diagnosis and procedure codes), were to be used. Employing such detailed coding schemes could enable the uncovering of new disease prognosis patterns, offering new opportunities for the predictive models to improve their representation learning and hence their prediction performance.  From this point, our discussion and subsequent analysis will focus only on models with matrix embeddings.

We have used \texttt{ICE-NODE\_UNIFORM} as a variant of \texttt{ICE-NODE} where all time-intervals between consecutive timestamps are fixed to one week, as a mean to evaluate within our framework the impact of incorporating the irregular intervals in the modeling, beyond using sequential information alone. Figure~ \ref{fig:relative_competency} shows that there is no clinical code predicted by \texttt{ICE-NODE\_UNIFORM} with a significantly higher \gls*{auc} than \texttt{ICE-NODE} in any of the three experiments (A-C), whereas \texttt{ICE-NODE} significantly outperforms \texttt{ICE-NODE\_UNIFORM} in 17 clinical codes in Experiment B and 7 clinical codes in Experiment C.
The results in Table~\ref{tab:top15acc} and Fig.~\ref{fig:relative_competency} show that the inclusion of temporal information (sequential only or full timestamps) is strongly advantageous, as seen by the reduced performance of \texttt{LogReg}. Furthermore, the analysis of Experiment B in Fig.~\ref{fig:auc_m4} shows that many clinical codes (18 out of 84) are predicted better by both \texttt{GRU} and \texttt{RETAIN} than by \texttt{ICE-NODE}. This suggests that for these clinical codes, the temporal-order information is sufficient without incorporating the irregular intervals, which, in turn, may have added noisy information in the learning of \texttt{ICE-NODE}, thus undermining its performance. 

The results for Experiment C (training on \gls*{mimic4} and predicting on \gls*{mimic3} as a test set) in relation to Experiment B (training and testing on splits of \gls*{mimic4}) in Fig.~\ref{fig:relative_competency} also suggest better generalisation properties for \texttt{ICE-NODE} across datasets. Specifically, \texttt{ICE-NODE} is competent in predicting 62 clinical codes in Experiment B, and this number is reduced by 12 in Experiment C when predicting across datasets. For \texttt{RETAIN} and \texttt{GRU}, this number is reduced by 27 and 30, respectively. 

Finally, we have shown that \texttt{ICE-NODE} produces time-continuous risk trajectories for each patient and each clinical code starting from the initial discharge from hospital. This feature may enable the use of the framework to gain data-informed insights to questions such as ``\textit{How will the risk of renal failure evolve within two weeks from now?}'' or ``\textit{Will the patient undergo an alarming risk of renal failure? and when will the risk exceed 0.5?}''.   \appendixref{appendix:trajectories} shows additional exemplars of such predicted trajectories.

\paragraph{Future Work} In relation to this last point, the user of \texttt{ICE-NODE} might be interested in studying time-continuous trajectories of multiple continuous simultaneously, to gain insights on how multiple medical conditions co-develop with time, and whether causal or correlational relationships might exist between conditions or other confounding conditions might explain these developments. This opens a future research direction to investigate whether the predicted, time-continuous trajectories can be reliable for causal discovery and/or causal inference tasks. This task becomes more challenging as the number of conditions studied increases but could help generate hypotheses on how medical conditions develop and interact supported by \texttt{ICE-NODE} data-based predictions. 

%




\paragraph{Limitations}

Using neural \glspl*{ode} can be sensitive to noisy information compared to the baseline models. This could explain why 18 out of 84 clinical codes in Experiment B are better predicted with two of the baseline models. The noise sources that can limit \texttt{ICE-NODE} learning capacity can be attributed to two reasons. First, for some clinical codes, the temporal-order could be sufficient for predictability, and incorporating the time-intervals is unnecessary and a source of variability. Second, mislabeled codes (false positives) and/or unreported codes (false negatives) in the ground-truth can misguide the prediction of the clinical codes trajectories. Since neural \glspl*{ode} integrate trajectories based on noisy initial conditions, the sensitivity to noise increases with a longer time-interval until the next hospital discharge. 
Another limitation in this work concerns the training itself. We have observed that training \texttt{ICE-NODE} improves remarkably when using two learning rates (one for the parameters of the dynamics and another for the rest of the parameters). This burdens the task of searching for optimal hyperparameters, hence further improvements for \texttt{ICE-NODE} could be possible.  At present, we do not have a sound explanation for this behaviour in the training.

When this research was finished, we learned of research by \citet{peng2021sequential} using neural \glspl*{ode}  to predict clinical codes of future visits. Even though that work exploits the temporal dimension, we discuss in \appendixref{appendix:related} their design which is pragmatically optimised for predicting future codes without producing time-continuous trajectories for the patient state, in contrast to our work.



\section{Conclusions}

We have presented a new disease progression model (\texttt{ICE-NODE}) that learns from timestamped clinical codes using neural \glspl*{ode}, and fully exploits the time dimension by incorporating the irregular time-intervals between the timestamps, as opposed to only exploiting the temporal-order. We have provided performance analyses by applying \texttt{ICE-NODE} (and the baselines) on \gls*{mimic3} and \gls*{mimic4}. These analyses have identified a set of clinical codes that are predicted with improved performance using \texttt{ICE-NODE}, such as `Ac renal fail' and `Pulm hart dx'. Our analyses also show better generalisation properties for \texttt{ICE-NODE} across datasets. We have finally discussed the implications of obtaining time-continuous risk trajectories for diseases, to improve the understanding of disease development and interactions. We have provided pointers for future directions of research on exploiting these trajectories to reveal potential causal relations between the medical conditions as they evolve with time.

\acks{This work was supported by the UKRI CDT in AI for Healthcare \url{https://ai4health.io} (Grant No. EP/S023283/1) and the NIHR Imperial Biomedical Research Centre (BRC), Imperial Clinical Analytics, Research and Evaluation (iCARE). MB acknowledges support from the EPSRC under grant EP/N014529/1 funding the EPSRC Centre for Mathematics of Precision Healthcare at Imperial.}

\bibliography{references}

\newpage

\appendix

\section{Consort Diagrams}\label{appendix:consort}

\begin{figure}[h!]
    \centering
    \includegraphics[width=.6\linewidth]{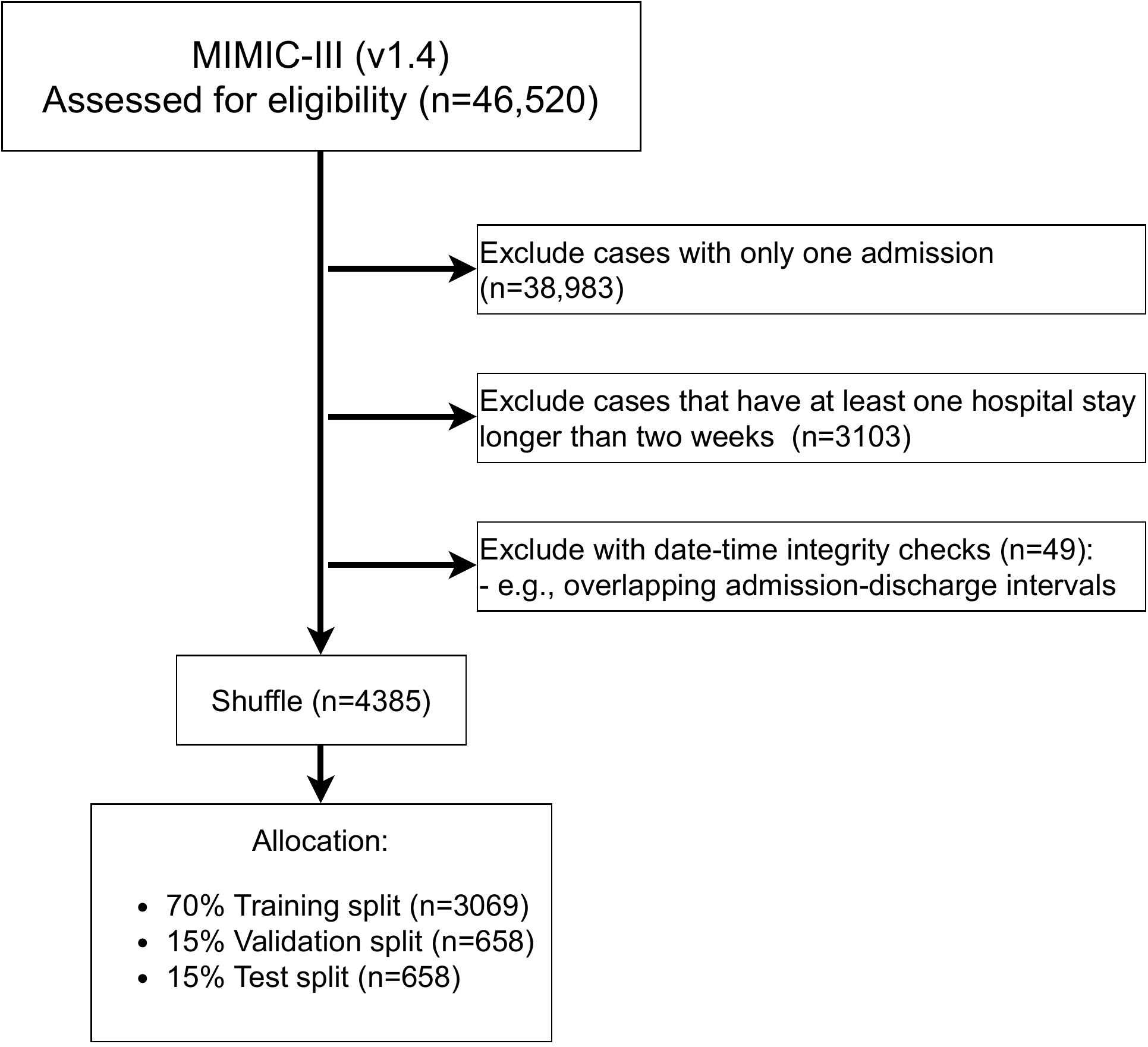}
    \caption{\footnotesize Consort diagram for the cohort extracted from \gls*{mimic3} for our experiments.}
    \label{fig:consort-mimic3}
\end{figure}

\begin{figure}[h!]
    \centering
    \includegraphics[width=.6\linewidth]{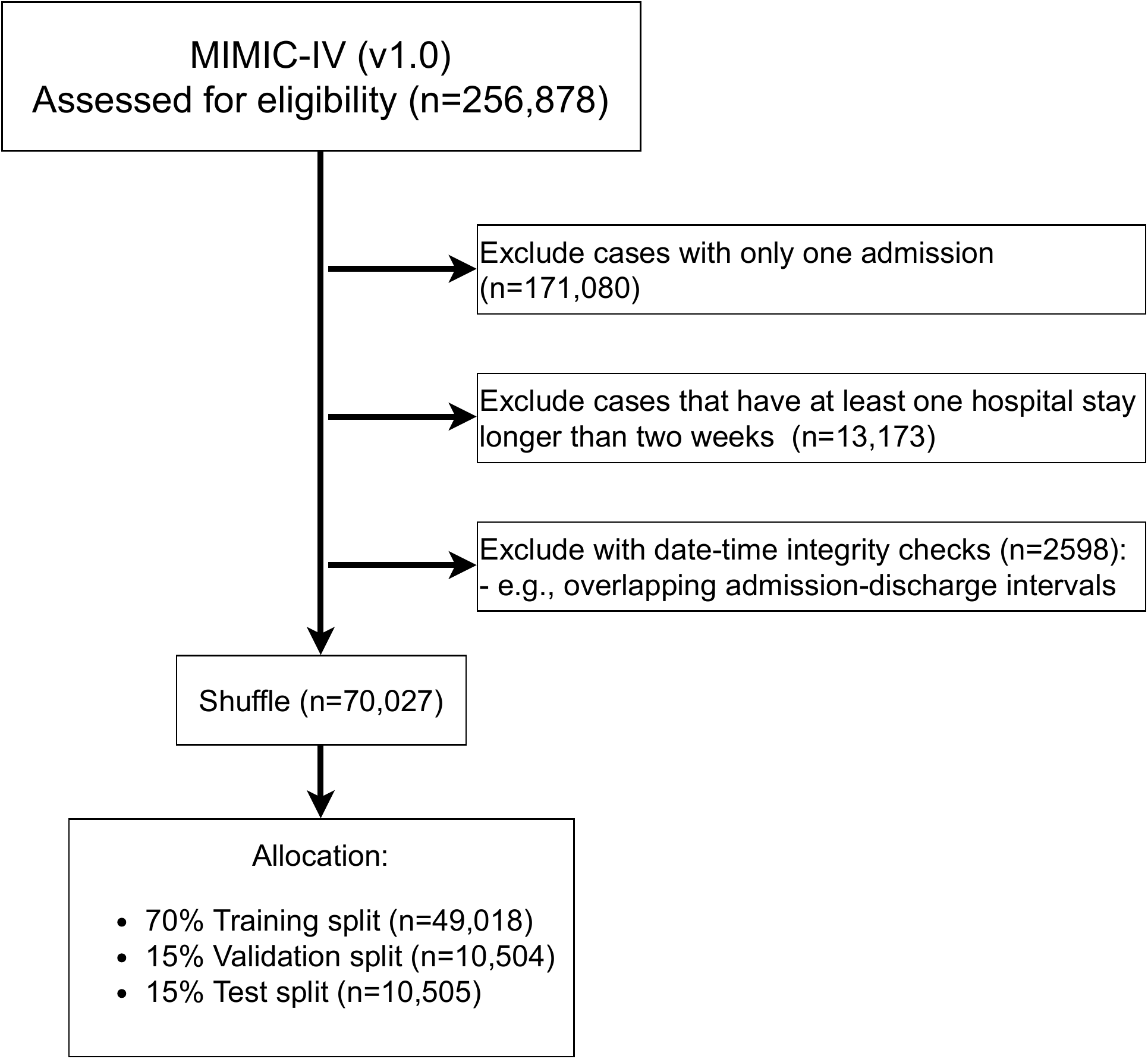}
    \caption{\footnotesize Consort diagram for the cohort extracted from \gls*{mimic4} for our experiments.}
    \label{fig:consort-mimic4}
\end{figure}

\newpage

\section{\texttt{ICE-NODE}: Implementation Details \& Optimal Configuration}
\label{appendix:config}

This section provides implementation details and description for the hyperparameter search task to find the optimal configuration of \texttt{ICE-NODE} and \texttt{ICE-NODE/G}, in addition to their corresponding variants \texttt{ICE-NODE\_UNIFORM} and \texttt{ICE-NODE\_UNIFORM/G}. However, the source code in \url{https://github.com/barahona-research-group/ICE-NODE} can be consulted for a better comprehension and the additional details of the optimal configuration finding for \texttt{GRU}, \texttt{GRU/G}, \texttt{RETAIN}, and \texttt{LogReg}. The implementation of our method and the baselines relies on the \emph{automatic differentiation} engine of \texttt{JAX} library \citep{jax2018github}, which also provides the scalable gradient computation of neural \glspl*{ode} using the \emph{adjoint method}~\citep{NEURIPS2018_69386f6b}.

We used the \texttt{optuna} library \citep{optuna_2019} to run the hyperparameter optimisation harnessing distributed runs across multiple nodes. We used the \emph{Tree-structured Parzen Estimator} algorithm \citep{NIPS2011_86e8f7ab} with the default parameters as in \texttt{optuna}\footnote{\url{https://optuna.readthedocs.io/en/stable/reference/generated/optuna.samplers.TPESampler.html}}.  We specified the averaged \emph{visit \gls*{auc}} metric on the validation partition of \gls*{mimic3} as the hyperparameter optimisation objective.

The hyperparameter space is divided into four categories: 
\begin{itemize}
    \item the clinical embeddings module,
    \item the dynamics function,
    \item the decoder function,
    \item and the training settings.
\end{itemize}    

Two main tasks are conducted: the first task is focused on optimising the configuration of \texttt{ICE-NODE} (with the matrix embeddings). The optimal configuration found for \texttt{ICE-NODE} is reused for \texttt{ICE-NODE\_UNIFORM}. The second task is focused on \texttt{ICE-NODE/G}, but the optimisation here is focused only on optimising the configuration of the \texttt{GRAM} embeddings, while the other configurations are fixed to the optimal configuration found for \texttt{ICE-NODE} in the first task. Similarly, the final optimal configuration of \texttt{ICE-NODE/G} is reused for \texttt{ICE-NODE\_UNIFORM/G}.

\subsection{The Clinical Embeddings Module}

\paragraph{Matrix Embeddings for \texttt{ICE-NODE}}

The only hyperparameter of the matrix embeddings is the embeddings vector size, i.e. $d_e$ in \equationref{eq:mat-emb}. The search domain of $d_e$ is $[30, 60, 90, \ldots, 300]$. The optimal value is found to be $d_e = 300$.

\paragraph{\texttt{GRAM} Embeddings for \texttt{ICE-NODE/G}}

The optimisation of the configuration of \texttt{GRAM} embeddings considered three hyperparameters: \\

\begin{tabular}{|l|c|c|c|}
    \hline
     hyperparameter & type & domain & optimal value \\
     \hline
     $d_e$: embedding dimension  & integer & $[30, 60, 90, \ldots, 300]$ & 300 \\
     attention method & categorical & [\texttt{tanh}\eqref{eq:f_tanh}, \texttt{l2}\eqref{eq:f_l2}] & \texttt{tanh}\\
     $\ell$: hidden layer dimension  & integer & $[50, 100, \ldots, 300]$ & 200 \\
     \hline
\end{tabular}\\

We use the \emph{GloVe} initialisation of the basic embeddings that has been described in \citep{Choi2017GRAM:Learning}.

\subsection{The Dynamics Function}

The configuration of the dynamics function considered two hyperparameters

\begin{tabular}{|l|c|c|c|}
    \hline
     hyperparameter & type & domain & optimal value \\
     \hline
     $d_m$: memory state size  & integer & $[10, 20, 30, \ldots, 100]$ & 30 \\
     architecture & categorical & [\texttt{mlp2}, \texttt{mlp3}, \texttt{gru}] & \texttt{mlp3}\\
     \hline
\end{tabular}\\

While each architecture is defined as following:
\begin{itemize}
    \item \texttt{mlp2}: is an \gls*{mlp} with two layers, where each layer has (i) the dimensionality $d_h=d_m + d_e$, (ii) the activation function $\tanh$, and (iii) no bias term.
    \item \texttt{mlp3}: is same as \texttt{mlp2} but with three layers.
    \item \texttt{gru}: is a minimal \gls*{gru} described in \citep[Appendix G]{Brouwer2019GRU-ODE-Bayes:Series}, and originally authored by \citet{ZhouWZZ16}.
\end{itemize}

\subsection{The Decoder Function}
The decoder function is implemented as a \gls*{mlp}, where:
 (i) hidden layers has the dimensionality of $d_e$ and followed by \texttt{LeakyReLU} activation, 
 while (ii) the output layer has the dimensionality $C$ (i.e. the coding scheme vocabulary size) and followed by \emph{sigmoid} activation.
The only hyperparameter here is the number of layers. The domain of search was limited to $[2, 3]$, and the optimal number of layers found is 2.

\subsection{The Update Function}

The update function of \texttt{ICE-NODE} in \equationref{eq:update} adjusts the memory state to accommodate the new information at the new timestamp, and it is implemented as following:
\begin{align*}
    \bm{h}_m(t_k^+) &= f_U(\bm{h}_m(t_k^-), \bm{h}_e(t_k^-),
    \bm{g}(t_1); \bm{\theta}_U) \\
    &= \texttt{GRU}(\bm{W}_U \begin{bmatrix}
           \bm{h}_m(t_k^-) \\
           \bm{g}(t_k)
         \end{bmatrix} + \bm{b}_U,  \bm{h}_m(t_k^-); \bm{\theta}_U),
\end{align*}
where $\bm{W}_U \in \mathbb{R}^{d_e \times 2d_e}$ and $\bm{b}_U \in \mathbb{R}^{d_e}$ updates the concatenation 
$[\bm{h}_m(t_k^-); \bm{g}(t_k)]$ 
to a space with the same dimensionality $R^{d_e}$.  $\texttt{GRU}: R^{d_e} \times R^{d_m} \mapsto R^{d_m}$  is an implementation of a \gls*{gru} cell \citep{cho2014learning}\footnote{\url{https://dm-haiku.readthedocs.io/en/latest/api.html\#gru}}, which computes the updated memory state. No new hyperparameter to tune for the update function, as the dimensions $d_e$ and $d_m$ are already included above.

\subsection{The Training settings}

For each training iteration, we randomly sample, with replacement, a fixed number of patients $B$ from the total of $N$ patients in the training partition. With a slight abuse of terminology, we use the term `epoch' to refer to a number of training iterations equals to $\lceil N/B\rceil$, where $\lceil . \rceil$ is the ceiling operator. The number of `epochs' is fixed to 60. The hyperparameters of the training settings is listed below:\\

\begin{tabular}{|l|c|c|c|}
    \hline
     hyperparameter & type & domain & optimal value \\
     \hline
     optimiser & categorical & [\texttt{adam}, \texttt{adamax}, \texttt{sgd}] & \texttt{adam} \\
     $\eta_1$: dynamics learning rate & float & \texttt{LogUniform}$[10^{-5}, 10^{-2}]$ & $7.15\times 10^{-5}$ \\
     $\eta_2$: the other learning rate & float & \texttt{LogUniform}$[10^{-5}, 10^{-2}]$ & $1.14\times 10^{-3}$ \\
     \texttt{decay\_rate} & float & \texttt{LogUniform}$[0.1, 0.9]$ & 0.3 \\ 
     $B$: batch size & integer & $[2, 4, 8, 16, \ldots, 256]$ & 256 \\
     \hline
\end{tabular}\\

\paragraph{Early stopping for model selection through the training iterations} Since we rely on iterative methods for training our models we adopt the \emph{early stopping}\footnote{\url{https://en.wikipedia.org/wiki/Early_stopping}} strategy to avoid overfitting our model after a large, fixed number of iterations. Throughout the training iterations, we evaluate the averaged \emph{visit \acrshort*{auc}} on the \glspl*{ehr} of the validation set. The \emph{visit \acrshort*{auc}} is evaluated for each visit after the first discharge in the patient history, and it estimates the probability of assigning risk scores to the actual clinical codes issued at that visit with higher risk values than those of the unreported clinical codes. This metric is used to guide the model selection throughout the training iterations. 

\newpage
\section{\gls*{auc} values for clinical codes highlighted in \figureref{fig:relative_competency}}\label{sec:auc_vals}

\begin{figure}[htbp]
\floatconts
{fig:auc_vals}
{\caption{\footnotesize
 \gls*{auc} values for clinical codes highlighted in red in \figureref{fig:relative_competency}. The 3 codes in (a) are detected equally well by all methods that use either sequence or full temporal information but not by \texttt{LogReg}, which ignores time altogether. The codes in (b) and (c) are only predicted well by \texttt{ICE-NODE}, but not by methods that use only sequence information or by \texttt{LogReg}.  }} %
{%
\subfigure[Experiment A][b]{
        \label{fig:subset_auc_m3}
        \includegraphics[width=0.4\textwidth]{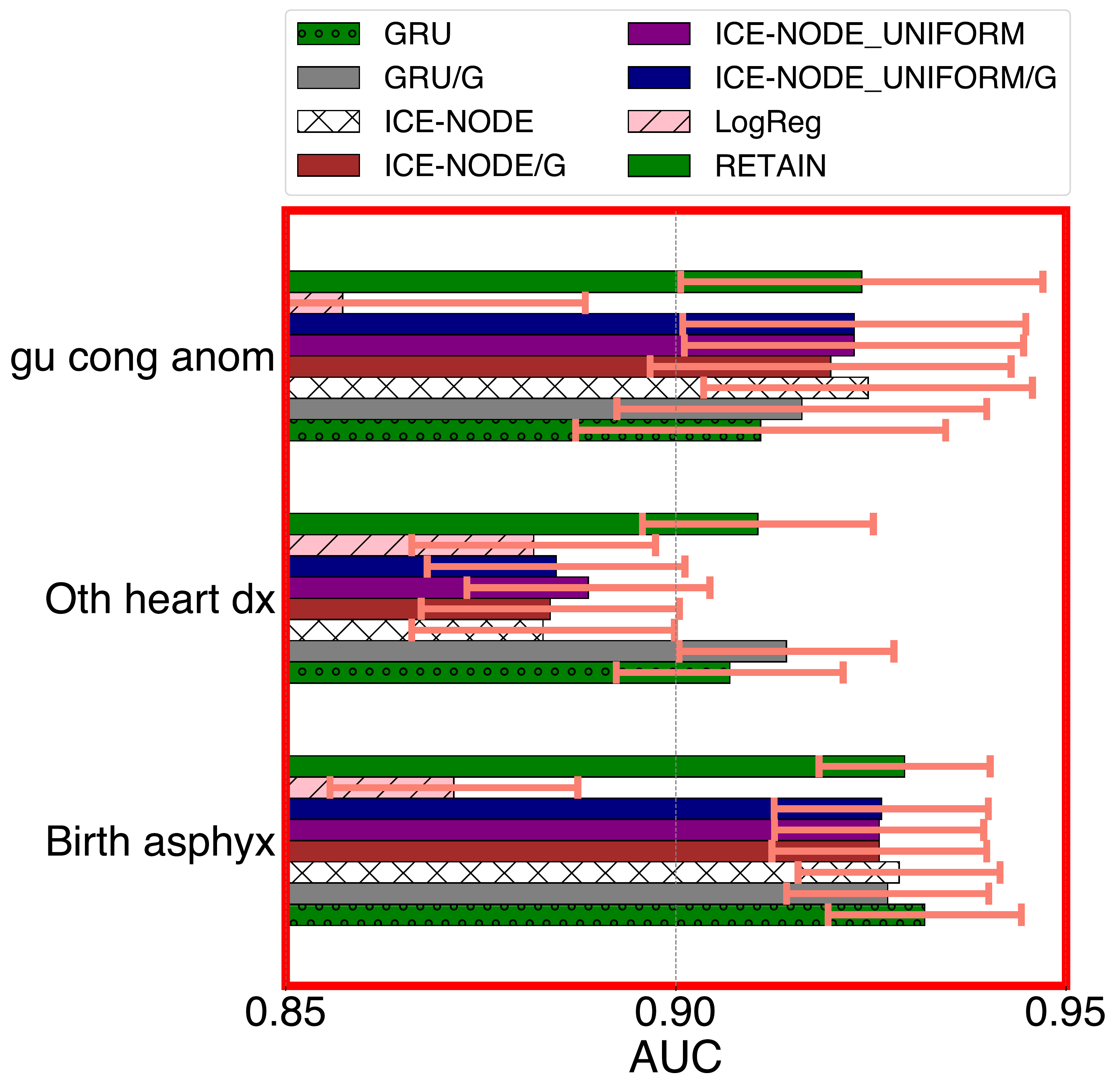}
}
\subfigure[Experiment B][b]{
        \label{fig:subset_auc_m4}
        \includegraphics[width=0.4\textwidth]{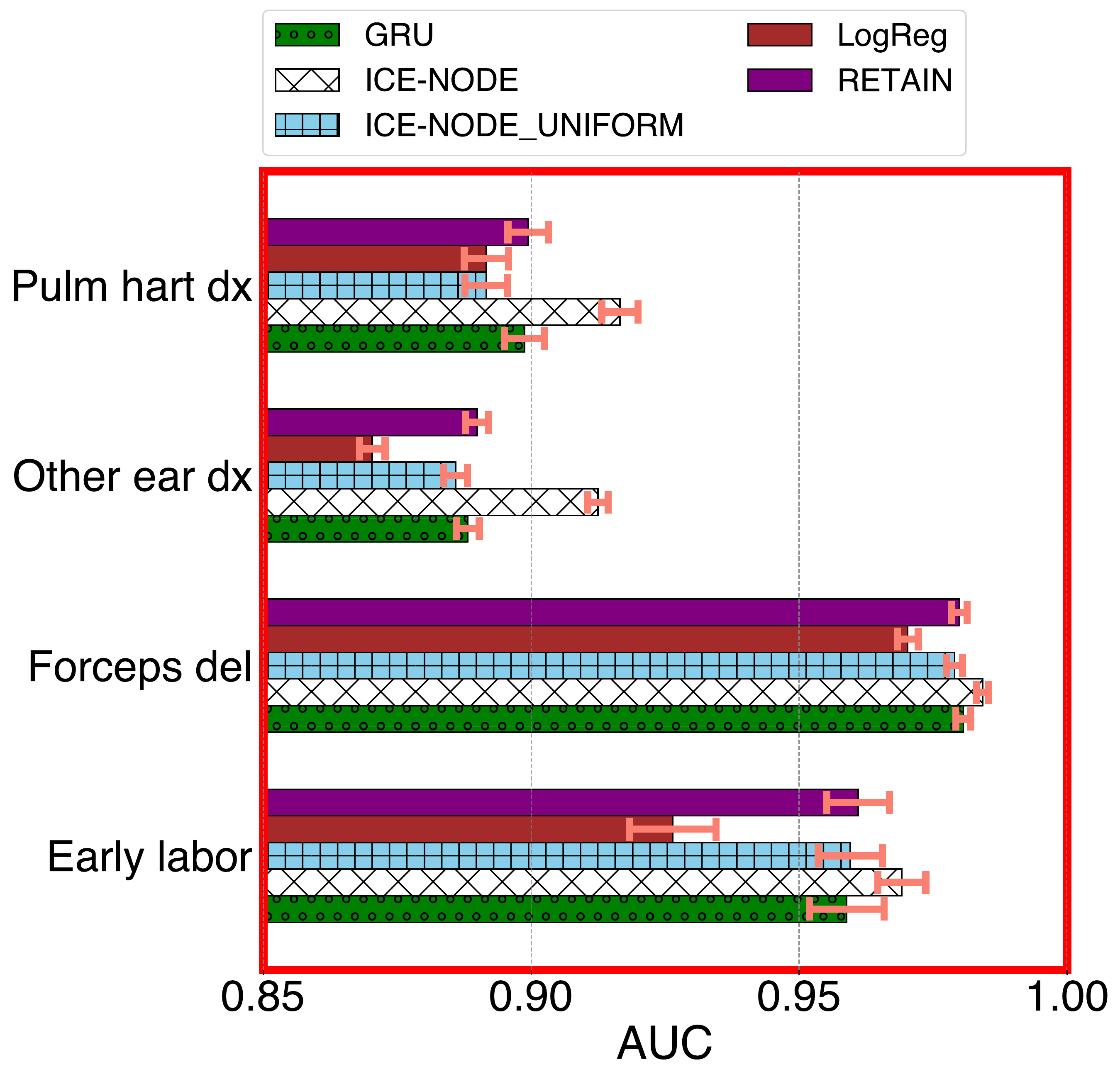}
}
\subfigure[Experiment C][b]{
        \label{fig:subset_auc_m4m3}
        \includegraphics[width=0.4\textwidth]{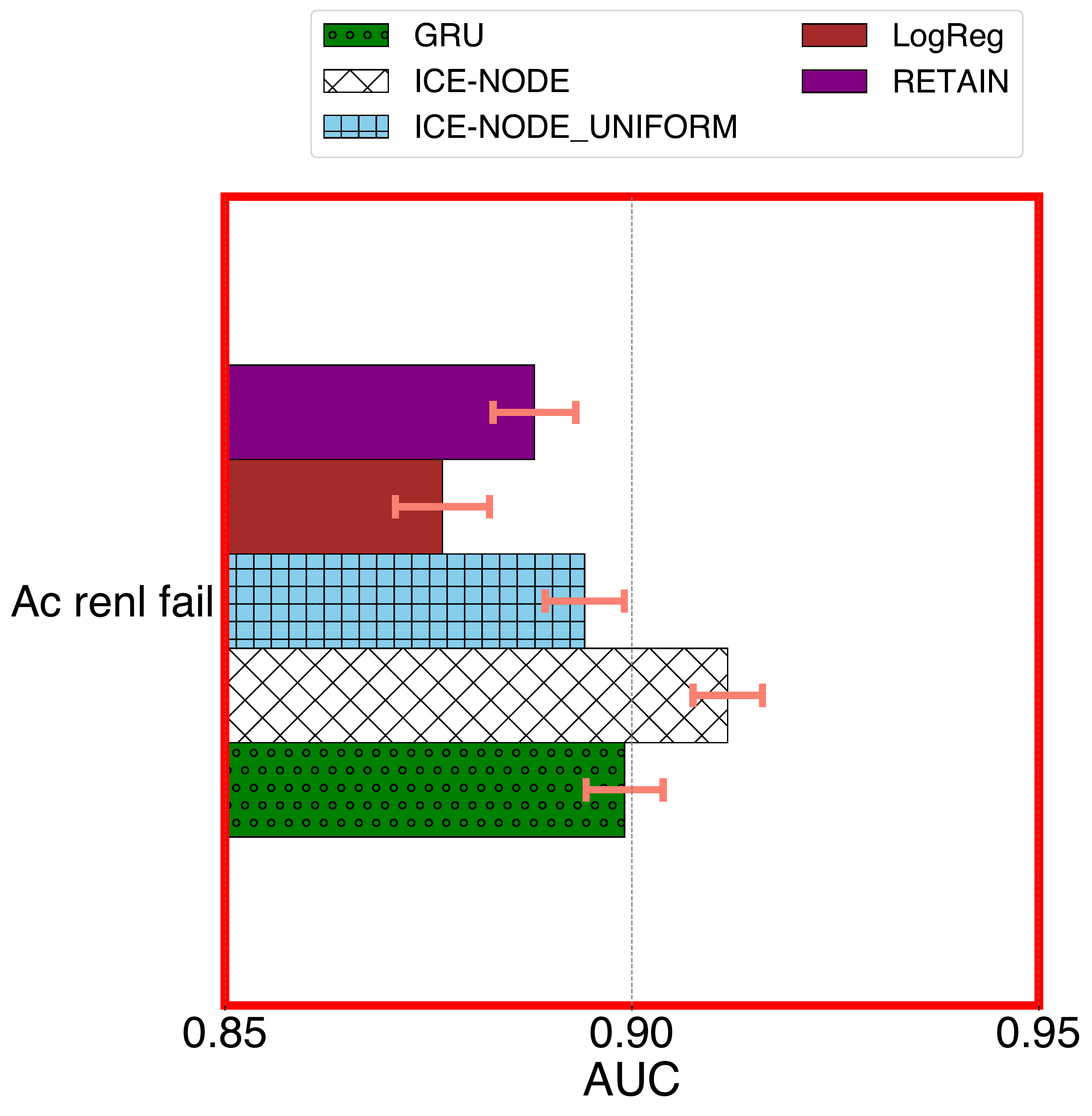}
}

}
\end{figure}

\newpage
\section{Further Examples of \texttt{ICE-NODE} Predicted Risk Trajectories}\label{appendix:trajectories}

\begin{figure}[htbp]
    \centering
    \includegraphics[width=.6\linewidth]{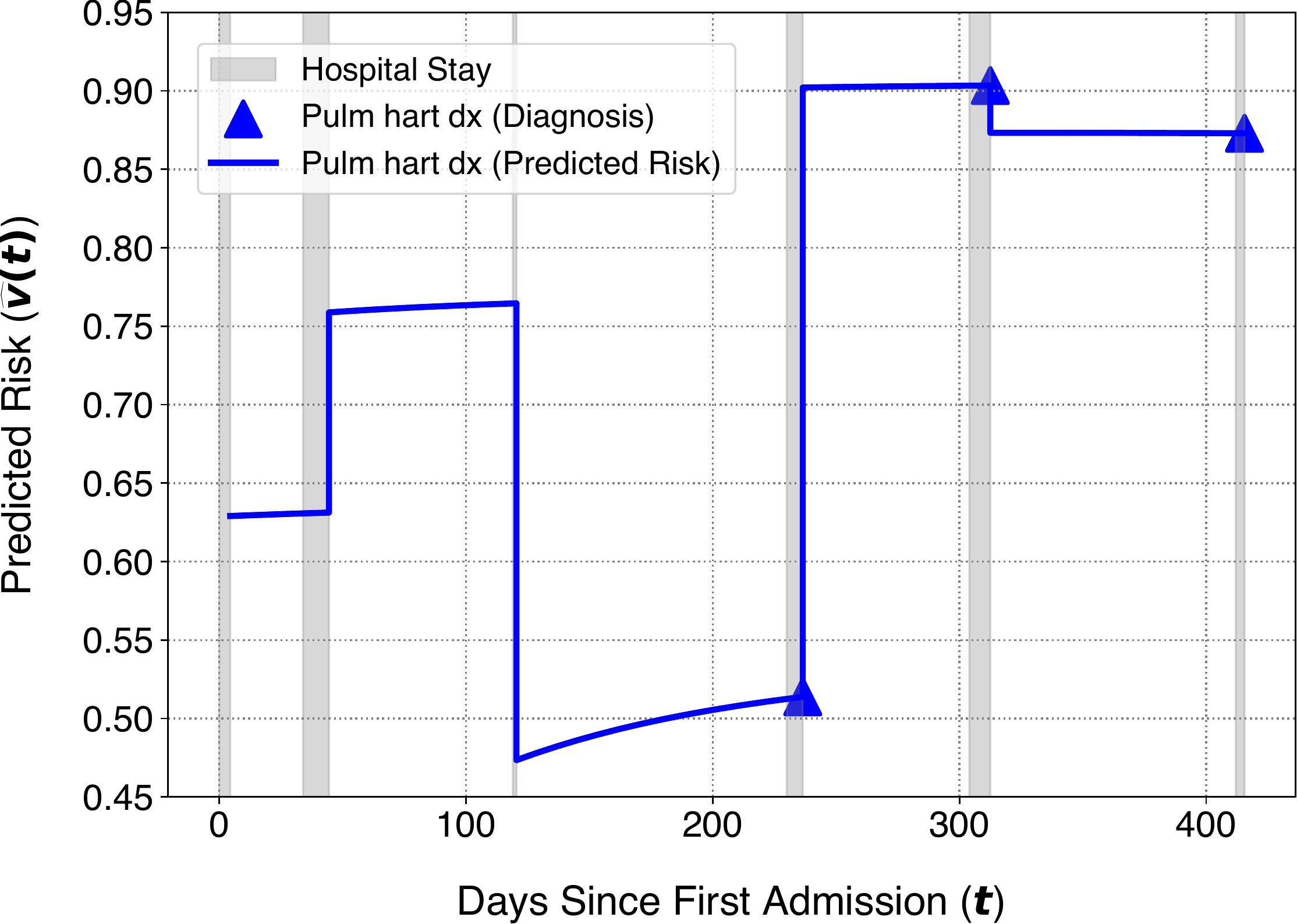}
    \caption{\footnotesize The predicted risk trajectory of `\emph{Pulm hart dx}' for the patient with \texttt{subject\_id=50093} in the test partition of \gls*{mimic3}. The history of this patient consists of six hospital stays (i.e. admissions-discharges) and the patient was diagnosed with `\emph{Pulm hart dx}' for the first time at the fourth hospital stay onward. After the initial discharge, \texttt{ICE-NODE} has predicted a risk for this diagnosis with a probability slightly lower than 0.65. The risk increased after the second discharge despite a negative reporting of `\emph{Pulm hart dx}', but decreased after another negative reporting at the third discharge, after which the risk alarmingly increased with time until the fourth hospital stay.}
    \label{fig:traj_m3_2}
\end{figure}


\begin{figure}[htbp]
    \centering
    \includegraphics[width=.6\linewidth]{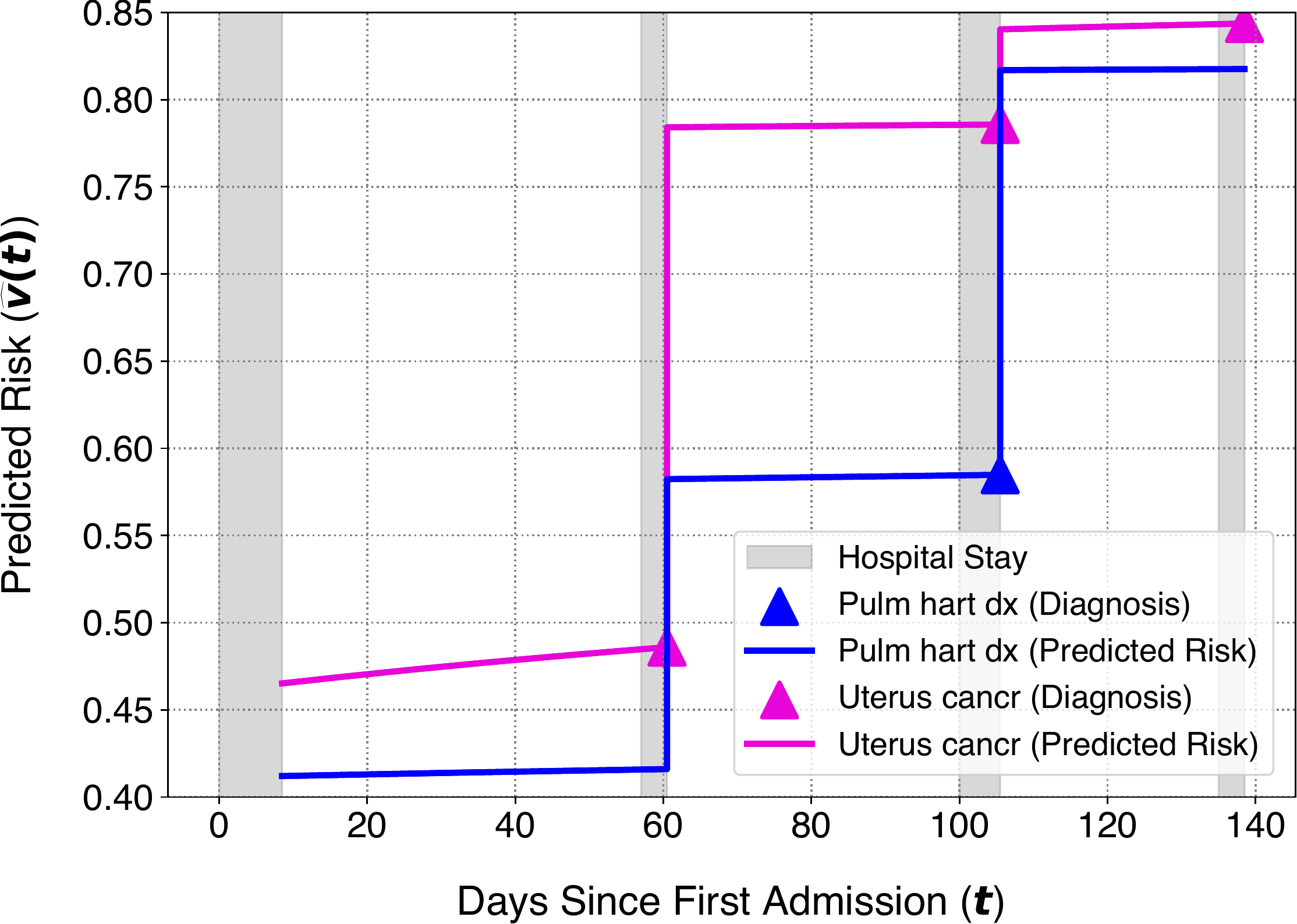}
    \caption{\footnotesize The predicted risk trajectory of `\emph{Pulm hart dx}' together with `\emph{Uterus cancr}' for the patient with \texttt{subject\_id=11052692} in the test partition of \gls*{mimic4}. The history of this patient consists of four hospital stays and the first diagnosis of  `\emph{Uterus cancr}' was made at the second hospital stay, while `\emph{Pulm hart dx}' at the third hospital stay.}
    \label{fig:traj_m4_2}
\end{figure}

\newpage
\section{Discussion on recent related work by \citet{peng2021sequential}}
\label{appendix:related}
When this research was finished, we learned that a solution based on neural \glspl*{ode} and transformers has been employed by \citet{peng2021sequential} to capture information about irregular intervals within the \glspl*{ehr}. 
Briefly, their approach aims to produce a representation $\bm{v}^o_t$ for each $t$-th visit. Using $f$ to denote multiple computational modules, their representation is generated as $$\bm{v}^o_t = f(\bm{v}^{\text{code}}_t, \bm{v}^{\text{LoS}}_t, \bm{v}^{\text{Interval}}_t)$$ 
where $\bm{v}^{\text{code}}_t$ is the clinical code representation of that visit; $\bm{v}^{\text{Interval}}_t$ is the final solution of a certain \gls*{ivp} from the discharge timestamp of the previous visit to the current admission timestamp,
and $\bm{v}^{\text{LoS}}_t$ (LoS is short for length-of-stay) is the final solution of a certain \gls*{ivp} from the visit admission timestamp until the discharge timestamp.
The visit representation $\bm{v}^o_t$ is then used to predict the clinical codes of the next visit using one-layer \emph{softmax}. 

Whereas this approach exploits the temporal dimension pragmatically to produce accurate predictions for next visits, it fuses three different representations, including the ones that capture the irregular intervals in the \gls*{ehr}.
Hence it lacks the advantage of producing intermediate representations spanning the time intervals between visits. In contrast, \texttt{ICE-NODE} can retrieve the patient state at any arbitrary time, allowing for the retrieval of time-continuous trajectories.

\end{document}